\documentclass[sn-mathphys,Numbered]{sn-jnl}

\usepackage{booktabs}%
\usepackage{graphicx}%
\usepackage{multirow}%
\usepackage{amsmath,amssymb,amsfonts}%
\usepackage{amsthm}%
\usepackage{mathrsfs}%
\usepackage[title]{appendix}%
\usepackage{xcolor}%
\usepackage{textcomp}%
\usepackage{manyfoot}%
\usepackage{algorithm}%
\usepackage{algorithmicx}%
\usepackage{algpseudocode}%
\usepackage{listings}%
\usepackage{makecell}%
\usepackage{array}
\usepackage{verbatim}
\usepackage{graphics}
\usepackage{enumerate}
\usepackage{geometry}
\usepackage{caption}
\usepackage{subfigure}
\usepackage{lineno}
\usepackage{float}
\usepackage{afterpage} 
\usepackage{titletoc}

\theoremstyle{thmstyleone}%
\theoremstyle{thmstyletwo}%

\theoremstyle{thmstylethree}%

\newcommand{\beginextendeddata}{%
\setcounter{table}{0}
\renewcommand{\tablename}{Extended Data Table}
\setcounter{figure}{0}
\renewcommand{\figurename}{Extended Data Fig.}
\setcounter{algorithm}{0}
\setcounter{section}{0}
}

\newcommand{\beginsupplement}{%
\setcounter{table}{0}
\renewcommand{\tablename}{Table}
\renewcommand{\thetable}{S\arabic{table}}%
\setcounter{figure}{0}
\renewcommand{\figurename}{Fig}
\renewcommand{\thefigure}{S\arabic{figure}}%
\setcounter{algorithm}{0}
\renewcommand{\thealgorithm}{S\arabic{algorithm}}%
\setcounter{section}{0}
\renewcommand{\thesection}{S\arabic{section}}%
\setcounter{subsection}{0}
\renewcommand{\thesubsection}{Part~\arabic{subsection}}%
}

\begin{document}
\title[Article Title]{Generation of 3D Molecules in Pockets via Language Model}

\author[1]{\fnm{Wei} \sur{Feng}}
\equalcont{These authors contributed equally to this work.}

\author[1]{\fnm{Lvwei} \sur{Wang}}
\equalcont{These authors contributed equally to this work.}

\author[1]{\fnm{Zaiyun} \sur{Lin}}
\equalcont{These authors contributed equally to this work.}

\author[1]{\fnm{Yanhao} \sur{Zhu}}
\equalcont{These authors contributed equally to this work.}

\author[1]{\fnm{Han} \sur{Wang}}
\author[1]{\fnm{Jianqiang} \sur{Dong}}
\author[1]{\fnm{Rong} \sur{Bai}}

\author[1]{\fnm{Huting} \sur{Wang}}
\author[1]{\fnm{Jielong} \sur{Zhou}}   
\author[2]{\fnm{Wei} \sur{Peng}}

\author*[1]{\fnm{Bo} \sur{Huang}}\email{huangbo@stonewise.cn;orcid.org/0000-0003-3822-9110}
\author*[1]{\fnm{Wenbiao} \sur{Zhou}}\email{zhouwenbiao@stonewise.cn;orcid.org/0000-0002-7168-3676}

\affil[1]{\orgname{Beijing StoneWise Technology Co Ltd.}, \orgaddress{\street{Haidian Street \#15}, \city{Beijing}, \postcode{100080}, \country{China}}}

\affil[2]{\orgdiv{Innovation Center for Pathogen Research}, \orgname{Guangzhou Laboratory}, \orgaddress{\city{Guangzhou}, \postcode{510320}, \country{China}}}

\abstract{Generative models for molecules based on sequential line notation (e.g. SMILES) or graph representation have attracted an increasing interest in the field of structure-based drug design, but they struggle to capture important 3D spatial interactions and often produce undesirable molecular structures. To address these challenges, we introduce Lingo3DMol, a pocket-based 3D molecule generation method that combines language models and geometric deep learning technology. A new molecular representation, fragment-based SMILES with local and global coordinates, was developed to assist the model in learning molecular topologies and atomic spatial positions. Additionally, we trained a separate noncovalent interaction predictor to provide essential binding pattern information for the generative model. Lingo3DMol can efficiently traverse drug-like chemical spaces, preventing the formation of unusual structures. The Directory of Useful Decoys–Enhanced (DUD-E) dataset was used for evaluation. Lingo3DMol outperformed state-of-the-art methods in terms of drug-likeness, synthetic accessibility, pocket binding mode, and molecule generation speed.}



\maketitle
\addtocontents{toc}{\setcounter{tocdepth}{-1}}

\section{Introduction}\label{sec1}
Structure-based drug design, which involves designing molecules that can specifically bind to a desired target protein, is a fundamental and challenging drug discovery task\cite{anderson2003process}. \textit{De novo} molecule generation using artificial intelligence (AI) has recently gained attention as a tool for drug discovery. Earlier molecular generative models relied on either molecular string representations\cite{bjerrum2017molecular,kusner2017grammar,segler2018generating,xu2021novo} or graph representations\cite{li2018learning,liu2018constrained,jin2018junction,shi2020graphaf}. However, both representations disregard 3D spatial interactions, rendering them suboptimal for target-aware molecule generation. The increase of 3D protein-ligand complex structures data\cite{francoeur2020three} and advances in geometric deep learning have paved the way for AI algorithms to directly design molecules with 3D binding poses\cite{skalic2019target,gebauer2019symmetry}. For example, methods using 3D convolutional neural networks (CNNs)\cite{ragoza2022generating}  are used to capture 3D inductive bias, but they still struggles to  convert atomic density grids into discrete molecules. 

Some studies\cite{luo20213d,liu2022generating,peng2022pocket2mol,li2021structure} proposed to represent pocket and molecule as 3D graphs and used graph neural networks (GNNs) for encoding and decoding. These GNN models use an autoregressive generation process which linearizes a molecule graph into a sequence of sampling decisions. Although these methods can generate molecules with 3D conformations, they share some common drawbacks: (a) problematic substructures: the generated molecules often contain problematic, non-drug-like, or not synthetic available substructures such as very large rings (rings containing seven or more atoms) and honeycomb-like arrays of parallel, juxtaposed rings; (b) problematic topology: the generated molecules often contain an excessive number of rings or none at all. Autoregressive sampling method has its inherent limitations. It can easily get stuck in local optima during the initial stages of molecule generation and may accumulate errors introduced at each step of the sampling process. For example, as mentioned by reference\cite{guan20233d}, although the model can accurately position the $n^\text{th}$ atom to create a benzene ring when the preceding $n-1$ carbon atoms are already in the same plane\cite{peng2022pocket2mol}, accurate placement of the initial atoms is often problematic because of insufficient context information, resulting in unrealistic fragments. 

In addition, some methods based on other technical routes were proposed for 3D molecule generation, such as those based on diffusion models\cite{satorras2021n,hoogeboom2022equivariant,guan20233d}. The representative method is TargetDiff\cite{guan20233d}, which employs a graph-based diffusion model for non-autoregressive molecule generation. Despite its efforts to avoid autoregressive method, it still generates a notable proportion of undesirable structures. This problem is possibly caused by the model’s relatively weak perception of molecular topology, which is associated with its weak ability to directly encode or predict bonds. Consequently, although TargetDiff achieved improved performance compared to earlier models, it still has room for improvement in metrics such as quantitative estimate of drug-likeness (QED)\cite{bickerton2012quantifying} and synthetic accessibility score (SAS)\cite{ertl2009estimation}, highlighting the urgency of confining the generated molecules to a drug-like chemical space\cite{polishchuk2013estimation}.

While graph-based 3D molecular generation methods have shown great potential recently, they still face difficulties in reproducing reference molecules on a given pocket without any information leakage, which is an important benchmark for evaluation. To address the abovementioned problems, we propose Lingo3DMol. First, we introduced a novel sequence encoding method for molecules, called fragment-based simplified molecular-input line-entry system\cite{weininger1988smiles} (FSMILES). FSMILES encodes the size of the ring in all ring tokens, providing additional contextual information for the autoregressive method and adopts ring-first traversal to achieve improved performance. Furthermore, we integrated local spherical\cite{gebauer2019symmetry} and global Euclidean coordinate systems into our model. Because bond lengths and bond angles in the ligand are essentially rigid\cite{corso2022diffdock}, directly predicting them is an easier task than predicting the Euclidean coordinates of the atoms. Combining of these two types of coordinates enables the model to consider a larger spatial context while maintaining accurate sub-structures. Moreover, noncovalent interactions (NCI)\cite{ding2022observing} and ligand-protein binding patterns were also considered during molecule generation by incorporating a separately trained NCI/Anchor predictor. We also used 3D molecule de-noising pre-training strategies similar to BART\cite{lewis2019bart} and Chemformer\cite{chemformer} to improve the generalization ability of the model. Our model was finetuned with data from PDBbind2020\cite{renxiao2004pdbbind}. Finally, we evaluated Lingo3DMol on the DUD-E dataset and compared it with state-of-the-art (SOTA) methods. Lingo3DMol outperformed existing methods on various metrics.

Our main contributions can be summarized as follows:
\begin{itemize}
\item  A novel FSMILES molecule representation that incorporates both local and global coordinates is introduced, enabling the generation of 3D molecules with reasonable 3D conformations and 2D topology.
\item A 3D molecule de-noising pre-training method and an independent NCI/Anchor model are developed to help overcome the problem of limited data and identify potential NCI binding sites.
\item The proposed method outperforms SOTA methods in terms of various metrics, including drug-likeness, synthetic accessibility, and pocket binding mode.
\end{itemize}
\section{Results and Discussion}\label{sec:results_and_discussion}
\subsection{Datasets and Baselines}\label{subsec5}
\textbf{The pre-training dataset.} It was derived from an in-house virtual compound library comprising structures of more than 20 million commercially accessible compounds which are typically used for the virtual screening of drug hit candidates. Low-energy conformers were generated for each molecule using ConfGen\cite{watts2010confgen}. To exclude molecules with low drug-likeness, we applied a filtering process that removed complex rings and retained molecules with less than three consecutive flexible bonds, resulting in 12 million molecules. In this study, any rings that do not fall into the categories of 5-membered or 6-membered rings, as well as fused 5-membered or 6-membered rings were considered as complex rings.

\textbf{The fine-tuning dataset.} This dataset was sourced from PDBbind (general set, version 2020)\cite{renxiao2004pdbbind}, using the DUD-E dataset\cite{mysinger2012directory} as homology filters. Specifically, we excluded proteins from the training set that exhibited more than 30\% homology to any target in DUD-E, as determined using MMseqs2\cite{mirdita2021fast}. This process resulted in the selection of 9,024 PDB IDs, which represented approximately 46\% of the protein-ligand PDB IDs in the PDBbind database. Within these selected PDB IDs, non-crystallographic symmetry related protein-ligand complex molecules within an asymmetric unit were considered individual samples. Additionally, samples were excluded from training if no NCIs were recognized between the ligand and the pocket by the Open Drug Discovery Toolkit (ODDT)\cite{Maciej2015oddt}. As a result, we obtained a total of 11,800 samples, which encompassed 8,201 PDB IDs (i.e. 42\% of protein-ligand PDB ID in PDBbind), in the fine-tuning dataset.

\textbf{The NCI training dataset.} The samples in this dataset were the same as the fine-tuing samples. The \textbf{NCI}s of the hydrogen bond, halogen bond, salt bridge and pi-pi stacking in the PDBbind were labeled using ODDT. The anchors were marked as the atoms in the pocket that are less than 4 $\AA$ away from any atom in the ligand. These labeled samples were used for the NCI/Anchor prediction model, averaging 4.1 NCI atoms and 32.1 anchor atoms per pocket sample.  

\textbf{The test dataset.} The models were evaluated mainly using the DUD-E dataset, which includes over 100 targets and an average of over 200 active ligands per target. This dataset spans diverse protein categories such as Kinase, Protease, GPCRs, and ion channels. More importantly, the  experimentally measured affinity have been reported for the active compounds in DUD-E. Hence, it allows us to compare generated molecules with active ligands for various protein targets. The target with PDB ID 2H7L in the DUD-E dataset was excluded as it is listed as an obsolete entry in PDB.

\textbf{Baselines.} Two state-of-the-art (SOTA) models, Pocket2Mol\cite{peng2022pocket2mol} and TargetDiff\cite{guan20233d}, were used as baselines in this study. Pocket2Mol is an autoregressive generative GNN model, and TargetDiff is a diffusion-based model. These two models were respectively obtained from their official GitHub repository. As mentioned in their original research papers, these two models were trained using the CrossDocked2020 dataset\cite{francoeur2020three}.

\subsection{Model Evaluation}
In our evaluation, we conducted a comparative analysis of Lingo3DMol with baseline methods. We propose to evaluate the generated molecules from mainly three perspectives: molecular geometry, molecular property distribution, and the binding mode within the pocket.

\subsubsection{Molecular Geometry}
The bond length distribution of the generated molecules was assessed using a methodology similar to the one employed in the TargetDiff study. Specifically, around 10,000 molecules were generated for each of the three models tested in the study. These molecules were generated for 100 targets in the CrossDocked2020 dataset. Then we compared the bond length distribution of the generated molecules with that of reference molecules, consisting of 100 ligands selected from the CrossDocked2020 dataset, as used in the TargetDiff study. Both our model and benchmark models exhibited favorable performance, as indicated by similar mean bond lengths compared to the reference molecules (Extended Data Table \ref{tab:bond_lengths}). 

\begin{figure}[!h]
  \centering
  \includegraphics[width=1.0\linewidth]{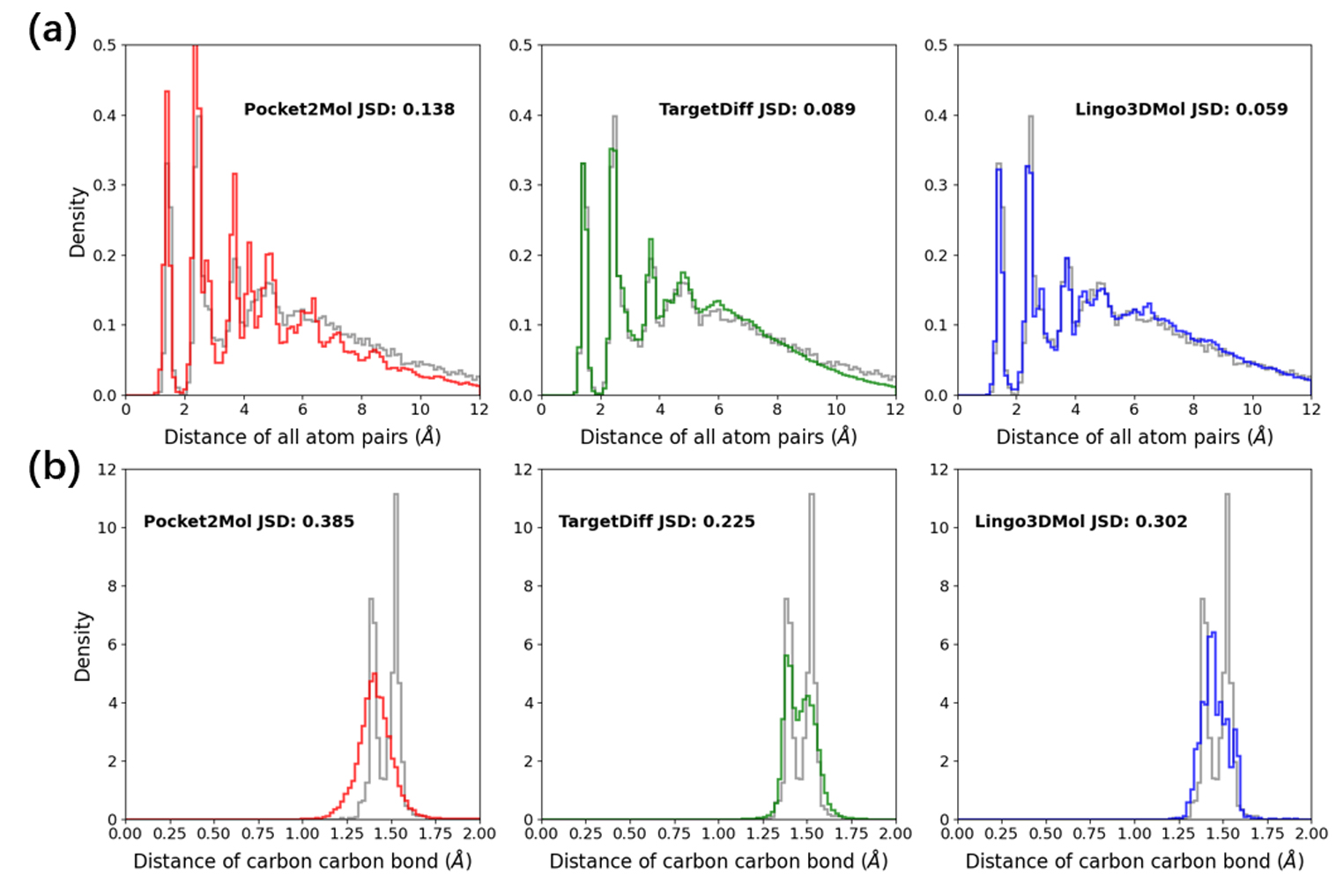}
  \caption{Comparison of atom-atom distance distributions in reference and generated molecules. (a) All atom pairs are considered in the analysis. (b) Only carbon-carbon atom pairs are considered.}
  \label{fig:atom_atom_dist}
\end{figure}

To assess the dissimilarity between the atom-atom distance distributions of the reference molecules and the model-generated molecules, we employed the Jensen-Shannon divergence (JSD) metric, which was also utilized in the TargetDiff study. Notably, Lingo3DMol demonstrated the lowest JSD score for all atom distances and ranked second for carbon-carbon bond distances (Extended Data Fig.\ref{fig:atom_atom_dist}).

Furthermore, we conducted an analysis of ring size, considering that molecules with large rings tend to be challenging to synthesize and may possess poor drug likeness. Our findings, as presented in Extended Data Table \ref{tab:ring_size}, revealed that our model exhibited a reduced tendency to generate molecules with a ring size of $>$7 compared to the TargetDiff and Pocket2Mol models. This observation suggests that our model shows promise in avoiding the generation of molecules with unfavorable ring size, further enhancing its potential for drug development applications.

\subsubsection{Molecular Property and Binding Mode}\label{ssub:binding_mode}
\textbf{Evaluation Metrics.} We proposed to test our model and baseline models using targets from DUD-E, because a significant number of experimentally measured active compounds were documented in this data set. Specifically, there are over 20,000 active compounds and their affinities against over 100 targets, an average of over 200 ligands per target. This allows us to analysis the similarity between generated molecules and known active compounds.

Regarding tools for binding pose evaluation, we propose to use Glide\cite{friesner2004glide} because it demonstrates superior ability in enriching active compounds and it is reported to be used as baseline in researches investigating scoring functions\cite{su2018comparative,shen2021beware}.

Regarding evaluation metrics for binding pose assessment, three options are available: min-in-place GlideSP score, in-place GlideSP score, and GlideSP redocking score. The min-in-place GlideSP score is obtained by using the "mininplace" docking method, where the ligand structure undergoes force field-based energy minimization within the receptor's field before scoring. It requires accurate initial placement of the ligand with respect to the receptor. The in-place GlideSP score is generated using the "inplace" docking method, which directly uses the input ligand coordinates for scoring without any docking or energy minimization. The GlideSP redocking score involves docking the generated molecules into the pocket, including the exploration of ligand binding conformations and initial placement.

Among the three docking-related metrics for binding pose evaluation, we advocate using the min-in-place GlideSP score for the following reasons. The in-place GlideSP score is excessively sensitive to atomic distances between the ligand and pocket, making it unsuitable for evaluating the quality of generated poses. In Extended Data Table \ref{tab:glidesp_anlysis}, a majority of molecules generated by all three methods (Lingo3DMol, Pocket2Mol, and TargetDiff) exhibit positive scores, indicating steric clashes between the pockets and ligands. However, such clashes do not necessarily denote a poor molecule unless they cannot be rectified by force-field-based minimization. On the other hand, the min-in-place GlideSP score respects the initial binding pose generated by the model and optimizes it through force-field-based adjustments to achieve a robust score. The use of only the GlideSP redocking score, without considering the min-in-place GlideSP score, would contradict the objective of 3D generation as it disregards the original pose. In this study, we recommend considering the min-in-place GlideSP score as the primary metric for binding pose evaluation, while also providing GlideSP redocking scores for contextual information.

\begin{figure}[!h]
  \centering
  \includegraphics[width=1.0\textwidth]{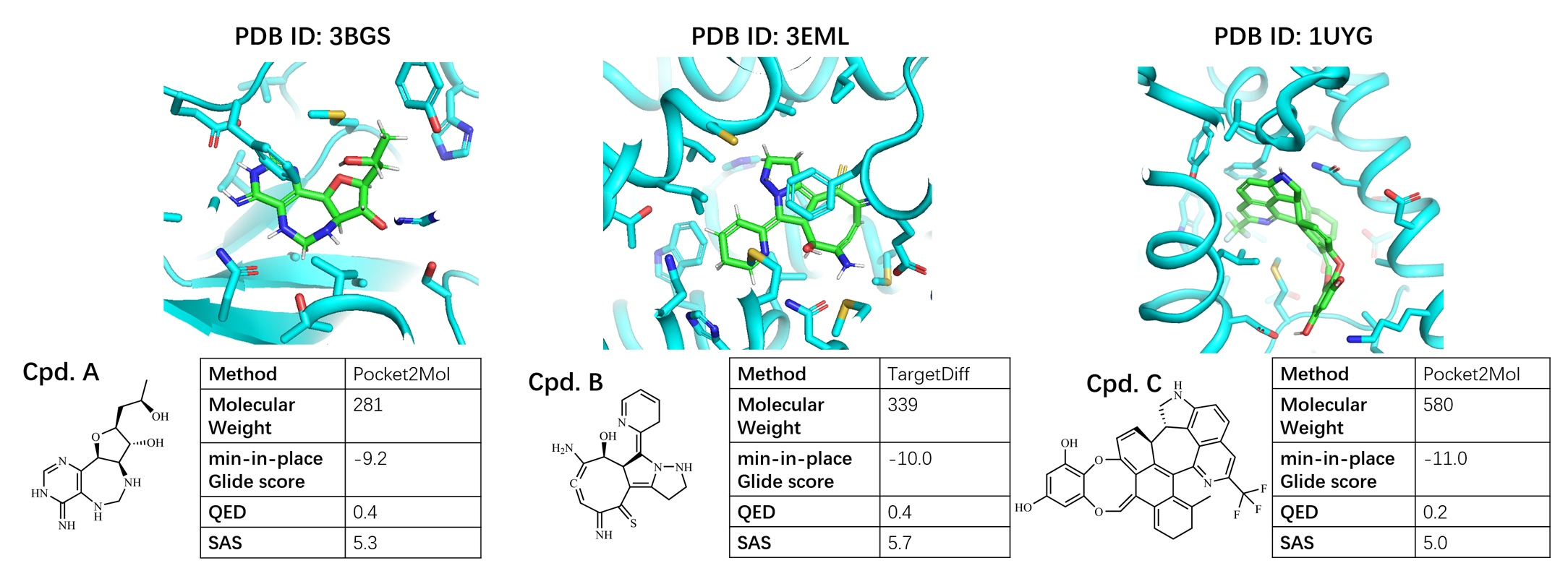}
  \caption{Cases of generated molecules with good min-in-place GlideSP scores but not suitable as drug molecules.}
  \label{fig:case_good_glidesp_bad_drug}
\end{figure}

\textbf{Molecular Property Distribution.} Prior to conducting the binding mode evaluation, we emphasize the significance of examining molecular property distributions. Extended Data Fig.\ref{fig:case_good_glidesp_bad_drug} shows three molecules that exhibit notably good docking scores (min-in-place GlideSP scores). However, despite their favorable scores, none of these molecules can be considered as potential drug candidates. Their exclusion stems from their poor drug likeness, as evidenced by low QED values, and low synthetic accessibility, reflected by high SAS values. This intriguing finding underscores the possibility that a superior performance on the binding pose evaluation metric may result from the presence of non-ideal molecules. Consequently, it becomes crucial to eliminate molecules with inadequate drug likeness or limited synthetic accessibility prior to calculating GlideSP scores. To further investigate the significance of this filtering criterion on a larger scale, we conducted an in-depth analysis of the distributions of various key properties of the generated molecules using heatmaps (Fig.\ref{fig:mol_dist}a to \ref{fig:mol_dist}e). 

It is important to notice that, as shown in the Fig.\ref{fig:mol_dist}c, the molecules that possess good min-in-place Glide scores (lower scores) are mostly found outside the drug-like region indicated by the red box for Pocket2mol and TargetDiff. To define the drug-like region, we considered a QED value of 0.3 or higher and a SAS value of 5 or lower, which encompass more than 80\% of the molecules in DrugBank\cite{10.1093/nar/gkx1037}. Unlike benchmark models, Lingo3DMol demonstrates a different pattern. Specifically, Lingo3DMol tends to generate drug-like molecules with relatively good min-in-place GlideSP scores. 

\textbf{Binding Mode Evaluation.}
Building upon the above analysis, we conducted a DUD-E-based evaluation of our model and the baseline models, with the results presented in Table \ref{tab:DUD-E}. Although the average QED and SAS values do not significantly differentiate our model from the baselines, the percentage of drug-like molecules determined by combining QED and SAS indicates the superiority of our model. 

Furthermore, in addition to generating drug-like molecules, an effective molecule generative model should be capable of generating active compounds. Since it is impractical to synthesize all generated molecules for real-world testing, an alternative approach is to evaluate whether the model can reproduce known active compounds or generate molecules that are highly similar to known active compounds. To assess this, we introduced the metric "ECFP\_TS$>$0.5", representing the percentage of targets with generated compounds that demonstrate over 0.5 Tanimoto similarity in terms of ECFP to active compounds. Among the drug-like molecules generated by the three models, our model yields similar-to-active compounds for 33\% of the targets, surpassing Pocket2Mol's 8\% and TargetDiff's 3\%. 

Additionally, for a 3D molecule generative model, it is crucial to generate molecules with favorable bindings in the target pockets. This assessment can be approached from two perspectives: binding mode with the pocket (interactions with the pocket) and the ligand's strain energy, both of which are typically associated with good binding affinity\cite{jain2023distributional,gu2021ligand,ryde2016ligand}. We utilized the min-in-place GlideSP score to evaluate pocket interactions and "RMSD vs. low-energy conformers" to reflect ligand strain energy. Although the "RMSD vs. low-energy conformers" metric does not directly quantify strain energy in the unit of kcal/mol, it provides valuable information on how closely the generated conformers resemble the low-energy conformers in terms of their overall geometry. This metric serves as a proxy for evaluating the ligand's strain energy. The generated conformations were compared against the top 20 lowest-energy conformers from ConfGen\cite{watts2010confgen}. Remarkably, Lingo3DMol outperforms the baselines in terms of the min-in-place GlideSP score and "RMSD vs. low-energy conformers", indicating the high quality of our generated conformations relative to the baselines. Moreover, our molecular generation speed is faster than benchmark methods (Extended Data Table \ref{tab:running_time}).

While our model exhibited slightly lower diversity in generating drug-like molecules compared to the baselines, it is important to note that the baselines' higher diversity does not translate into a satisfactory ability to generate similar-to-active compounds. This observation suggests that the baselines may explore the chemical space in a direction that deviates from the region where known active compounds are typically found. 

Lastly, we employed Dice score to access the 3D shape similarity between the reference compound observed in the crystal structure and the generated 3D molecules. By voxelizing the molecules and comparing their intersected and union points, we quantified the "intersection over union" ratio as Dice score which ranges from 0 to 1, with 0 indicating no similarity. Although all the models demonstrated similar performance according to this metric (as seen in Table \ref{tab:DUD-E}), the Dice score contributed in improving our model throughout the model development process (further details are provided in the Section \ref{sub:ablation}).

\textbf{Information Leakage Analysis.} Another significant point to consider is the issue of information leakage in the baselines during above DUD-E test. It is essential to note that Pocket2Mol was trained on the CrossDocked2020 dataset, which did not exclude targets with high homology to DUD-E targets. As a result, the performance of Pocket2Mol in this test may be overestimated due to the information leakage problem. On the other hand, our model was trained on a low-homology ($<$30\% sequency identity) dataset to mitigate this issue.

\begin{figure}[!h]
  \centering
  \includegraphics[width=1.0\textwidth]{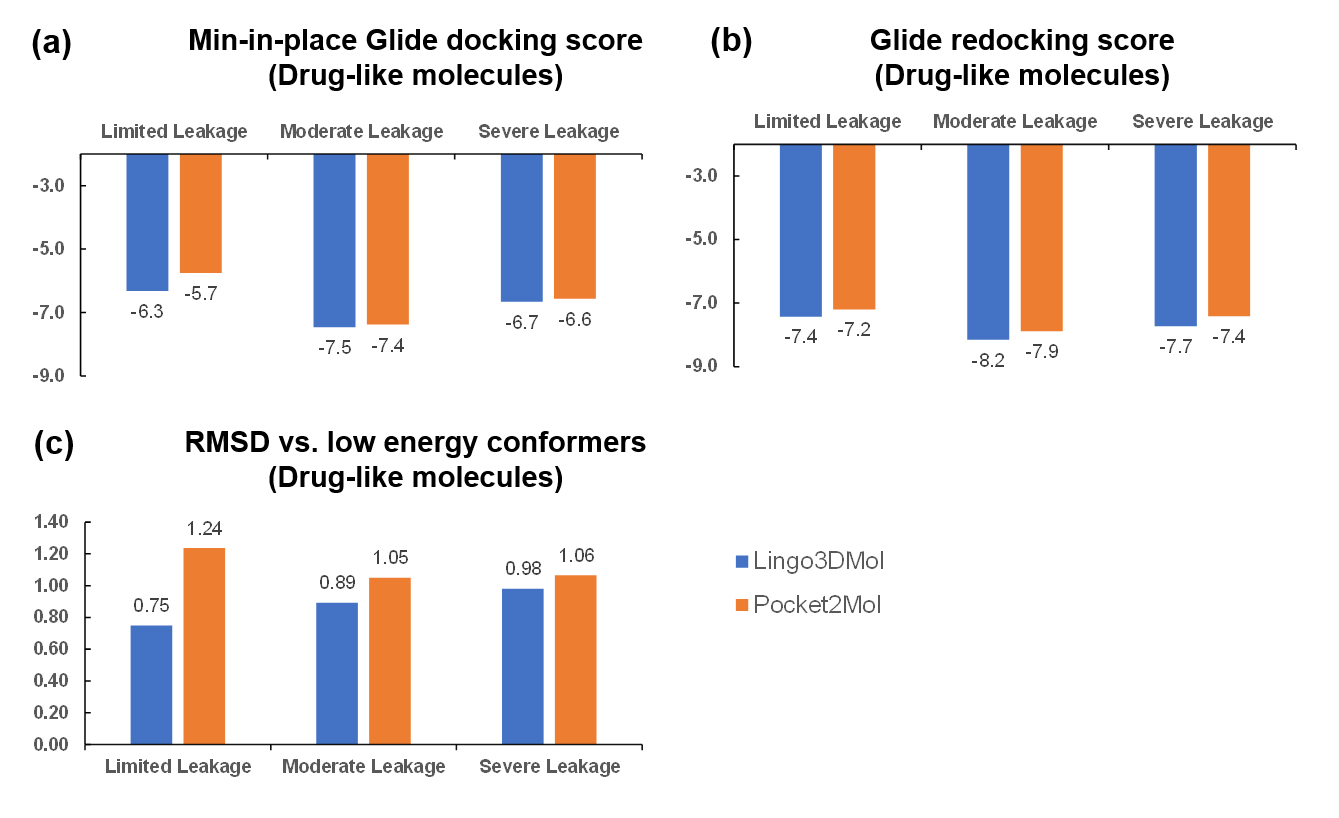}
  \caption{Comparison between Lingo3DMol and Pocket2Mol on the DUD-E dataset under varying degrees of information leakage. Lingo3DMol has limited information leakage by excluding proteins that have more than 30\% sequence identity with DUD-E targets from their training set. The assessment of information leakage is done from the perspective of Pocket2Mol. Specifically, DUD-E targets were categorized into three groups based on their sequency identity with Pocket2Mol training targets: severe ($>$90\%, N = 74), moderate (30-90\%, N = 19), and limited ($<$30\%, N = 8) information leakage. The comparisons on average min-in-place GlideSP scores, GlideSP redocking scores, and RMSD vs. low energy conformers are listed in panel (a), (b), and (c), respectively.}
  \label{fig:rmsd}
\end{figure}

To ensure fair comparisons, we categorized DUD-E targets into three groups based on their sequency identity with the Pocket2Mol training targets: severe ($>$90\%), moderate (30-90\%), and limited ($<$30\%) information leakage. Across all categories, Lingo3DMol consistently outperformed Pocket2Mol, particularly in terms of the min-in-place GlideSP score, Glide redocking score, and RMSD vs. low-energy conformers (Extended Data Fig.\ref{fig:rmsd}). It is intriguing to observe that the performance gap between the two models widens as the impact of information leakage in the baselines becomes insignificant.

\subsection{Case Analysis}\label{sub:case_anlysis}
For the case study, we selected the generated molecules from two perspectives: ECFP fingerprint similarity with known active compounds and the docking score. Particularly, a high similarity of ECFP fingerprints with known active compounds indicated the model’s capability of generating similar substructures or topology features with positive molecules, and a good docking score indicated a stronger fit with the pocket.

As shown in Fig.\ref{fig:my_fig}a and \ref{fig:my_fig}b, the molecules in "high similarity, good docking score" group resembled to positive molecules in terms of structure and binding mode, demonstrating the ability of our model to reproduce active compounds. However, this is insufficient for this study, because drug design researchers in the real world are more interested in retrieving the following two types of molecules: (a) active compounds that the docking program or other virtual screen tools fail to detect, and (b) molecules with novel scaffolds and good pocket binding affinity. The first issue, "active compounds missed by the docking program", is often caused by insufficient sampling of the binding pose, which is closely related to the docking score. Our 3D molecule generation model provides a potential solution to this issue. As shown in Fig.\ref{fig:my_fig}c and \ref{fig:my_fig}d, the generated molecules in the "high similarity, poor docking score" group are highly similar to the positive molecules but had poor docking scores (i.e. -6.5 and -6.4, respectively) when docking program was used for binding pose sampling (i.e. Glide redocking). Conversely, when a binding pose generated by our model was evaluated using Glide scorer without conformation sampling, we obtained good scores of -8.7 and -8.8 (i.e. min-in-place Glide score), respectively for the two compounds. This demonstrates the effectiveness of our generated 3D conformation for retrieving good molecules with poor docking scores. For the second issue, "obtaining molecules with a novel scaffold and good binding mode," we present two cases, as shown in Fig.\ref{fig:my_fig}e and \ref{fig:my_fig}f to demonstrate that our model can potentially generate molecules with these characteristics.

\subsection{Ablation Analysis}\label{sub:ablation}
\subsubsection{Effective Pre-training and Fine-tuning Analysis}\label{ssub:pretrain_and_finetuning}
Specifically, for DUD-E targets, the molecules generated by the models with and without pre-training were respectively compared with the molecules in the pre-training set. We demonstrated that the molecules generated by the pre-training model exhibited a higher degree of similarity to the molecules in the pre-training set compared to those generated by the model without pre-training. This indicates that the model retained the effect of pre-training after the fine-tuning. The comparison of these methods is described in Supplementary Information \ref{pre_train_supl}. As shown in Table \ref{tab:ablation_test}, pre-training significantly improves the percentage of drug-like molecules, mean QED, the percentage of ECFP\_TS$>$0.5, mean min-in-place GlideSP score and diversity. We attribute this improvement to the effectiveness of pre-training, especially in scenarios with limited fine-tuning data. In deep learning models, pre-training plays a crucial role in capturing relevant chemical patterns and features, allowing the model to generalize and generate molecules that align with desired properties even when fine-tuning data is limited.

\subsubsection{NCI Prediction Model Ablation Studies}\label{ssub:nci_ablation}
During this ablation study, we compared Lingo3DMol using randomly selected NCI sites to the standard Lingo3DMol which uses a well-trained NCI site predictor. It is important to note that both approaches share the same molecule generation model. As shown in Table \ref{tab:ablation_test}, standard Lingo3DMol demonstrated superior performance in most of the metrics, especially in drug likeness and ECFP\_TS$>$0.5. This can be attributed to several factors. One factor is that randomly selected NCI sites may result in the selection of solvent-exposed regions of the pocket where polar groups are more likely to be located. This may offer more accessible space compared to the cavity where the reference molecule is located. Additionally, the random selection of NCI sites tends to result in NCIs that are spaced farther apart from each other. The combination of these factors, including the preference for solvent-exposed regions and the spacing of selected NCIs, may contribute to the generation of larger molecules and subsequently impact the QED score and the percentage of drug-like molecules. 

Last but not least, it is worth noting that for over 95\% of DUD-E targets, both our training set (PDBbind, general set, version 2020) and the benchmark models' training set (CrossDocked2020) include at least one molecule with Tanimoto similarity greater than 0.5 to the DUD-E actives in terms of ECFP4 fingerprints. However, the significant improvement in ECFP\_TS$>$0.5 for standard Lingo3DMol compared to Lingo3DMol with random NCI and baseline models suggests that this improvement cannot be solely attributed to the model reproducing what it has seen during training.

\section{Conclusion}\label{sec:conclusion}
In this study, we proposed a novel molecule representation called FSMILES and developed Lingo3DMol, a model based on language modeling and geometric deep learning techniques. Compared with baselines, our model exhibited superior performance in generating drug-like 3D molecules with better binding mode. This indicates the potential of our model for further exploration in drug discovery and design.

Nevertheless, challenges remain. Capturing all non-covalent interactions within a single molecule is not straightforward due to the autoregressive generation process, and we plan to investigate this issue further. Representing molecules and inter-molecular interactions with electron densities perhaps offers a promising direction, and some related researches may serve as a good starting point\cite{wang2022pocket,ma2023using,ding2022observing}. Moreover, the equivariance property is a critical aspect of 3D molecule generation\cite{pmlr-v162-hoogeboom22a,satorras2021n,xu2022geodiff}. There are many works on rotational and translational equivariant models, such as GVP\cite{jing2021equivariant} and Vector Neurons\cite{deng2021vector}. Currently, we use rotation and translation augmentation to enhance the model, and we employ SE(3) invariant features like distance matrices and local coordinates\cite{simm2020symmetry} to alleviate the problem. Lastly, we have assessed drug-like properties through case analysis and employed cheminformatics tools such as QED\cite{bickerton2012quantifying} and SAS\cite{ertl2009estimation} from RDKit\cite{landrum2016rdkit}. However, a comprehensive and systematic evaluation of these properties is an essential next step for further research.

\section{Methods}\label{sec3}
In this section, we present an overview of the Lingo3DMol architecture and its key attributes. The methodology comprises two models: the generation model, which is the central component, and the NCI/Anchor prediction model, an essential auxiliary module. These models share the same Transformer-based architecture. The overall architecture and pre-training strategies are illustrated in Fig.\ref{fig:3DMG}a to 3c. In the following text, unless specifically mentioned, model is generally referring to the generation model.

\subsection{Definations and Notations}\label{def_note}
The Lingo3DMol learns $M \sim P(M|\text{Pocket}; \mu)$, where $\mu$ is the parameter of the model, $\text{Pocket}=(p_1,p_2...p_n)$ is the set of atoms in the pocket, and \(p_i = (\text{type}_i, \text{main/side}_i, \text{residue}_i, \text{coords}_i, \text{hbd/hba}_i, \text{NCI/Anchor}_i)\) indicates the information of the $i^\text{th}$ atom in the pocket, where "type" denotes the element type of the atom, "main/side" denotes an atom on the main/side chain, "residue" denotes the residue type of the atom, "coords" is the coordinates of the atom, "hbd/hba" denotes whether an element is a hydrogen bond donor or acceptor, and "NCI/Anchor" records whether it is a possible NCI site or anchor point where a potential ligand atom exists within a 4 $\AA$ range. Further details are provided in the Section \ref{NCI and anchor Predictor}. $M = (\text{FSMILES}, \{(r_{i})\}^{K}_{i=1})$ is the representation of the ligand, and $r_{i}$ is the coordinates of the $i^\text{th}$ atom of the ligand, $K$ is the number of atoms in the ligand.

\begin{figure}[!h]
  \centering
  \includegraphics[width=1.0\linewidth]{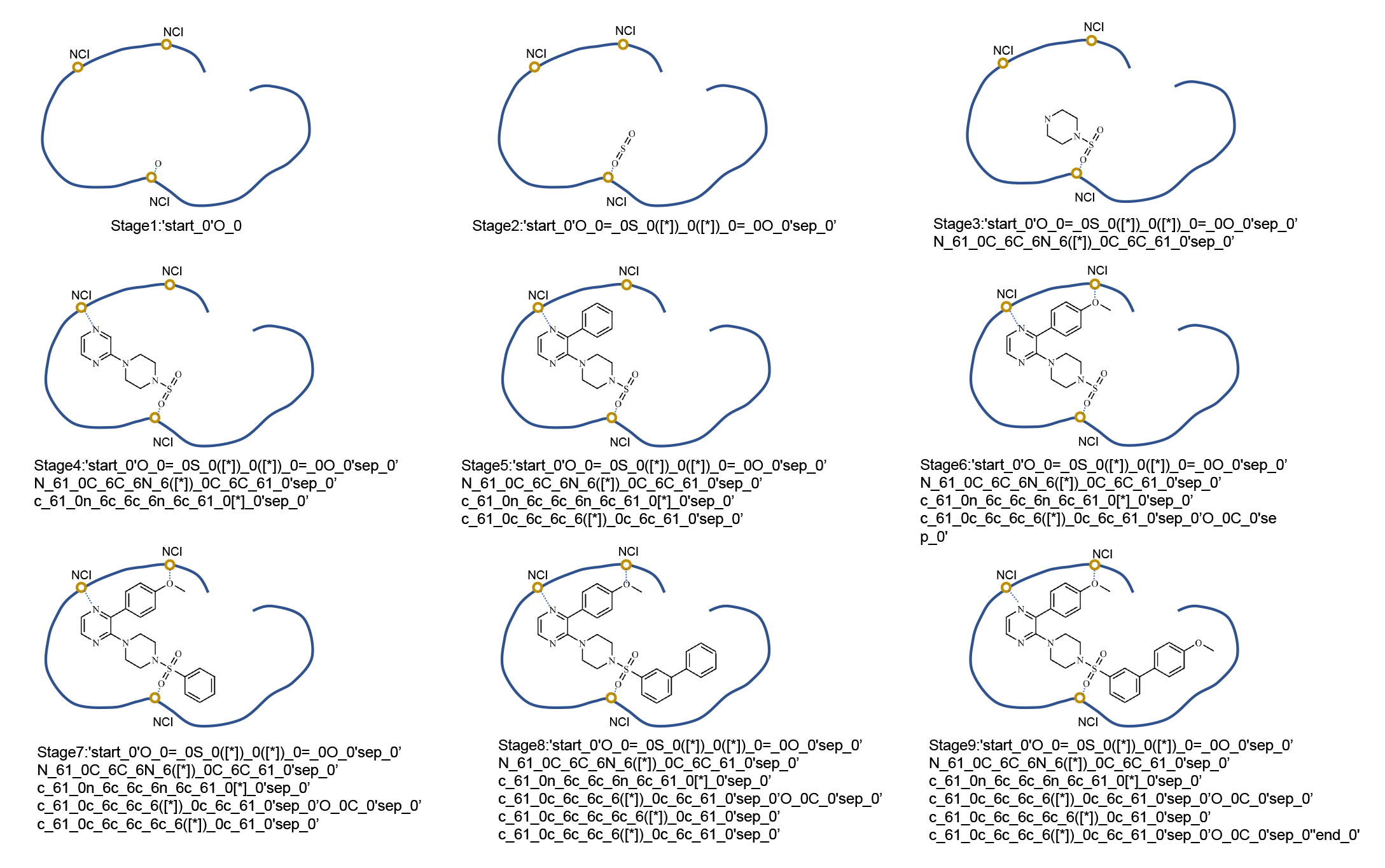}
  \caption{FSMILES based molecule growing process. Nine steps are listed to illustrate the fragment-by-fragment process of generating a molecule within a pocket.}
  \label{fig:main_process}
\end{figure}

\textbf{FSMILES}. FSMILES is a modified representation of SMILES\cite{weininger1988smiles} that reorganizes the molecule into fragments, utilizing the normal SMILES syntax for each fragment. The entire molecule is then constructed by combining these fragments using a specific syntax in a fragment first then depth-first manner, as illustrated in Extended Data Fig.\ref{fig:main_process}. This approach offers two key advantages: enhanced 2D pattern learning through the use of symbols to represent fragments and local structures, and the prioritization of ring closure, enabling the generation of molecules with accurate ring structures and bond angles. In FSMILES, the size of a ring is indicated from the first atom of the ring. For example, "C\_6" represents a carbon atom in a 6-membered ring. By providing both the atom type and the ring size in advance, the model can more accurately predict the correct bond angles.

The molecule fragment cutting process in FSMILES involves selecting individual bonds that meet specific criteria, such as not being part of a ring, not connecting hydrogen atoms, and having at least one end attached to a ring. This cutting process helps divide the ligand into fragments. The FSMILES construction process occurs in two steps (Fig.\ref{fig:3DMG}b). First, the ligand is divided into fragments according to the cutting rule. Second, ring information is embedded in each FSMILES token, with the number following the element type's underscore indicating the ring size. The symbol "*" denotes the connection points of a fragment, while the preceding atom indicates the connection position. In the depth-first growth model, each subsequent fragment connects to the preceding asterisk. To facilitate this, asterisks are stored in a stack, and when encountering a new fragment, the topmost asterisk is used to establish a connection.

\subsection{Pre-training and Fine-tuning}\label{subsec3}
\subsubsection{Pre-training Strategy}
In the pre-training phase, as illustrated in Fig.\ref{fig:3DMG}c, we introduced perturbations into the 3D molecular structure and fed the perturbed molecule into the encoder. This model, which employs an autoregressive approach, aims to reconstruct the perturbed molecule back to its original state in both 2D and 3D representations. 

\subsubsection{Fine-tuning}
For the fine-tuning phase, we utilized the pre-trained model and further fine-tuned it on the protein-ligand complex data. The primary task during this phase continued to be autoregressive molecule generation. To circumvent over-fitting of the fine-tuning dataset, the initial three encoder layers were fixed during fine-tuning.

The pseudocode of the above training process is shown in Supplementary Information Algorithm \ref{alg:training_process}.

\subsection{Model Architecture}
The generation model and NCI/Anchor prediction model were built upon the Transformer-based structure with additional graph structural information encoded into the model similar to the previous study\cite{ying2021transformers}. The generation model was trained by pre-training and fine-tuning. The NCI/Anchor prediction model was trained based on the generation model's pre-trained parameters and fine-tuned additionally by its specific prediction task. The overall architecture is shown in Fig.\ref{fig:3DMG}a. In the following sections, we first discuss the encoder and decoder components of the generation model, followed by an introduction to the NCI/Anchor prediction model.

\subsubsection{Encoder}\label{ssub:encoder_part}

\textbf{Input of Encoder during Pre-training Stage.} In the pre-training stage, the input of the encoder is a perturbed 3D molecule, which includes the element type and Euclidean coordinates. We can define the molecule as $M^{\text{enc}}=(m^{\text{enc}}_1,m^{\text{enc}}_2...m^{\text{enc}}_n)$,  \(m_i = (\text{type}_i, \text{coords}_i)\), where \(\text{coords}_i = (x_i, y_i, z_i)\), and $n$ is the number of atoms. Input feature $f_{\text{pre}}$ can then be defined as follows:

\begin{equation}\label{coords}
f_{\text{coords},i} = \text{MLP}([E_{\text{coords}}(x_i) , E_{\text{coords}}(y_i), E_{\text{coords}}(z_i)]),
\end{equation}

\begin{equation}\label{pretrain_input}
f_{\text{pre},i} = E_{\text{type}}(\text{type}_i) + f_{\text{coords},i},
\end{equation}
where
\begin{equation*}
\begin{split}
&E_{\text{type}}(\text{type}_i) \in \mathbb{R}^{H},f_{\text{coords},i}\in \mathbb{R}^{H}\\
&E_{\text{coords}}(x_i) \in \mathbb{R}^{H}, \ E_{\text{coords}}(y_i) \in \mathbb{R}^{H}, \ E_{\text{coords}}(z_i) \in \mathbb{R}^{H},
\end{split}
\end{equation*}
where $E_{\text{type}}$ and $E_{\text{coords}}$ represent the embedding functions for the element type, and the corresponding coordinates, respectively. $H$ is the size of the embedded vectors. The symbol "+" represents the element-wise addition operator, and the symbol "[$\cdot$]" represents the concatenation operator.

\textbf{Input of Encoder during Fine-tuning Stage.} In the fine-tuning stage, the input is changed from perturbed molecules to pockets. The input feature $E_{\text{fine}}$ can be defined as follows:

\begin{equation}
\begin{split}
f_{\text{fine},i} = & E_{\text{type}}(\text{type}_i) + E_{\text{main/side}}(\text{main/side}_i) + E_{\text{residue}}(\text{residue}_i) \\
& + f_{\text{coords},i} + E_{\text{hbd/hba}}(\text{hbd/hba}_i)+ E_{\text{NCI/Anchor}}(\text{NCI/Anchor}_i)
\end{split}
\end{equation}
where 
\begin{equation*}
\begin{split}
&E_{\text{main/side}}(\text{main/side}_i) \in \mathbb{R}^H,
E_{\text{residue}}(\text{residue}_i) \in \mathbb{R}^H,\\
& E_{\text{hbd/hba}}(\text{hbd/hba}_i) \in \mathbb{R}^H,
E_{\text{NCI/Anchor}}(\text{NCI/Anchor}_i) \in \mathbb{R}^{H}
\end{split}
\end{equation*}
where \(E_{\text{main/side}}\), \(E_{\text{residue}}\), \(E_{\text{hbd/hba}}\), and \(E_{\text{NCI/Anchor}}\) represent the embedding functions for the main/side chain, residue type, hydrogen bond donor/acceptor, and NCI/Anchor point, respectively.

\subsubsection{Decoder}
The molecule generation process is implemented by two decoders: one 2D topology decoder ($D_{\text{2D}}$) generates FSMILES tokens and local coordinates, and the other decoder generates 3D global coordinates ($D_{\text{3D}}$). First, the next token is predicted using $D_{\text{2D}}$, then the latest 2D token is input to $D_{\text{3D}}$. The 3D global coordinates decoder predicts the global coordinates of the new fragment in the molecule. Both decoders simultaneously predict the local coordinates $r$, $\theta$, and $\phi$, particularly the radial distance ($r$), bond angle ($\theta$), and dihedral angle ($\phi$) of the molecule. The local coordinates prediction by $D_{\text{3D}}$ only serves as an auxiliary training task.

\textbf{Input of Decoder.}
The input of the decoder network for a molecule can be defined as $M^{\text{dec}} = (m^{\text{dec}}_1,m^{\text{dec}}_2...m^{\text{dec}}_n)$,  \(m_i = (\text{token}_i, \text{global\_coords}_i)\) and  \(M_{\text{bias}} = (D, J)\), where \(D\) is a distance matrix of size \(n \times n\) and \(J\) is an edge vector matrix of size \(n \times n\), where \(n\) is the length of the sequence. $M^{\text{dec}}$ is transformed into embeddings using the same process as Equation \ref{pretrain_input}. These embeddings are then fed into the 2D topology decoder and global coordinates decoder, respectively. For non-atom tokens, we assigned the same coordinates as those of the most recently generated atom. 

\textbf{Attention Bias.}
Bias terms $B_D$ and $B_J$ are derived from the distance and edge vector matrices $D$ and $J$, respectively. Modified attention scores were calculated by incorporating the following bias terms:

\begin{equation}
    A_\text{biased} = \text{softmax}\left(\frac{QK^\top }{\sqrt{d_k}}+ B_D + B_J\right)V.
\end{equation}

\textbf{Predicting FSMILES and Local Coordinates.}
For FSMILES, we used a MLP projection head to predict the next token based on $D_{\text{2D}}$'s output. To predict the local coordinates, we established a local coordinates system with three atomic reference points: $\text{root1}$ is current position's parent atom, root1's parent atom is $\text{root2}$, and $\text{root2}$'s parent atom is $\text{root3}$. Parent atom is the atom that connects to the child atom. 

To predict the radial distance $r$, we utilized features of $\text{root1}$ ($h_\text{root1}$) extracted from the $D_{\text{2D}}$'s hidden representation, the current FSMILES token ($E_{\text{type}}(\text{cur})$), and the molecule's hidden representation ($H_\text{topo}$) from $D_{\text{2D}}$, $H_{\text{topo}} = (h_1,h_2 ... h_i)$, where each $h$ represents a FSMILES token's hidden representation, and $i$ represents the number of the generated tokens. For the polar angle $\theta$, we concatenated the hidden representations $h_\text{root1}$ and $h_\text{root2}$. To predict the dihedral angle $\phi$, we concatenated the representations of all three roots.

We used MLPs as projection heads to predict the local coordinates $(r, \theta, \phi)$. The mathematical representations of these processes are as follows:

distance prediction $r$:
\begin{equation}
r = \text{argmax} \left( \text{softmax} \left( \text{MLP}_1 \left( [E_{\text{type}}(\text{cur}), H_\text{topo}, h_\text{root1}] \right) \right) \right),
\end{equation}

angle prediction $\theta$:
\begin{equation}
\theta = \text{argmax} \left( \text{softmax} \left( \text{MLP}_2 \left( [E_{\text{type}}(\text{cur}), H_\text{topo}, h_\text{root1}, h_\text{root2}] \right) \right) \right),
\end{equation}

dihedral angle prediction $\phi$:

\begin{equation}
\phi = \text{argmax} \left( \text{softmax} \left( \text{MLP}_3 \left( [E_{\text{type}}(\text{cur}), H_\text{topo}, h_\text{root1}, h_\text{root2}, h_\text{root3}] \right) \right) \right).
\end{equation}

The predicted local coordinates $(r, \theta, \phi)$ are obtained by taking the argmax operator of the corresponding softmax output.

\textbf{Predicting Global Coordinates}. As illustrated in Fig.\ref{fig:3DMG}a, $D_{\text{3D}}$ receives the hidden representation of $D_{\text{2D}}$, and concatenate the predicted FSMILES token, and then predicts the global coordinate $(x, y, z)$. 

\textbf{Pre-training/Fine-tuning Loss}. In our model, the loss function is a combination of multiple components that evaluates different aspects of the predicted molecule. The overall loss function is defined as follows:
\begin{equation}\label{loss}
L = L_\text{FSMILES} + L_\text{abs\_coord} + L_{r} + L_{\theta} + L_{\phi} + L_{r\_{\text{aux}}} + L_{\theta\_{\text{aux}}} + L_{\phi\_{\text{aux}}},
\end{equation}
where:
\begin{itemize}
    \item $L_{\text{FSMILES}}$ measures the discrepancy between the predicted and ground-truth molecular topology.
    \item $L_{\text{abs\_coord}}$ evaluates the difference between the predicted and ground-truth atomic coordinates.
    \item $L_{r}$ and $L_{r\_\text{aux}}$ measure the error between the predicted and ground-truth radial distances.
    \item $L_{\theta}$ and $L_{\theta\_\text{aux}}$ assess the discrepancy between the predicted and ground-truth bond angles.
    \item $L_{\phi}$ and $L_{\phi\_\text{aux}}$ evaluate the difference between the predicted and ground-truth dihedral angles.
\end{itemize} 

All the prediction tasks are treated as classification tasks. Therefore, the cross entropy loss is used for each individual loss component. Auxiliary prediction tasks $r\_\text{aux},\theta\_\text{aux}$, and $\phi\_\text{aux}$ are used only during training. They are not utilized during the actual inference process. For further details, please refer to the Section \ref{ssub:generation_process}.

\subsubsection{NCI/Anchor Prediction Model}\label{NCI and anchor Predictor}
In this work, we aimed to enhance the generation model by integrating NCI and anchor point information during fine-tuning and inference stage. To achieve this, we employed an NCI/Anchor prediction model, which mirrors the generation model's architecture. This prediction model was initialized using the generation model's pre-trained parameters. Equiped with a specialized output head, the encoder can predict whether a pocket atom will form an NCI with the ligand or act as an anchor point.

This approach allowed us to enhance the generation model in two significant ways. Firstly, we enriched the model's input  by incorporating the predicted NCI and anchor point data as distinct features of the pocket atoms; Secondly, we started the molecule generation process by predicting the first atom of the molecule near a chosen NCI site. Specifically, we sampled an NCI site within a pocket and generated the first small molecule atom within a 4.5 $\AA$ radius of this site.

There are three main implications of our approach:

\begin{itemize}
\item We can enhance the perception of NCI and pocket shape which is critical for generating 3D molecules that can effectively interact with the target protein.

\item By positioning the first atom near the atoms within the NCI pocket, we can increase the likelihood of obtaining a correct NCI pair with a high degree of certainty. Although our model is designed to generate molecules that are prone to forming NCI pairs, it cannot ensure the exact positioning of the generated molecule in relation to the NCI pocket atoms. This is because the model does not explicitly enforce the coverage of all predicted NCI pocket positions, thereby allowing for some degree of variability in the positioning of the generated molecule.

\item Obtaining a good starting position: the autoregressive generation can be seen as a sequential decision-making problem. If the initial step is not chosen appropriately, it can affect every subsequent step. Our empirical study suggests that the NCI position as a starting point is a better choice than a random sample from the predicted 3D coordinates distribution or by selecting the coordinates with highest possibility.
\end{itemize}

\textbf{NCI/Anchor Prediction Loss}.
We defined two loss functions: The NCI loss measures the difference between the predicted NCI sites and the ground truth NCI sites, and the anchor loss measures the difference between the predicted anchor points and the ground truth anchor points. Both loss functions use binary cross entropy.

The total loss for the model is the sum of the NCI loss, anchor loss, and all other auxiliary losses from the original 3D generation task.

\begin{equation}
 L_{\text{total}} = L_{\text{NCI}} + L_{\text{Anchor}} + L_{\text{gen}},   
\end{equation}
where $L_{\text{NCI}}$ and $L_{\text{Anchor}}$ represent the NCI and anchor losses, respectively. $L_{\text{gen}}$ corresponds to the losses from the original 3D generation task which served as only an auxiliary training objective.

\subsection{Generation Process}\label{ssub:generation_process}
\begin{figure}[!h]
  \centering
  \includegraphics[width=1.0\linewidth]{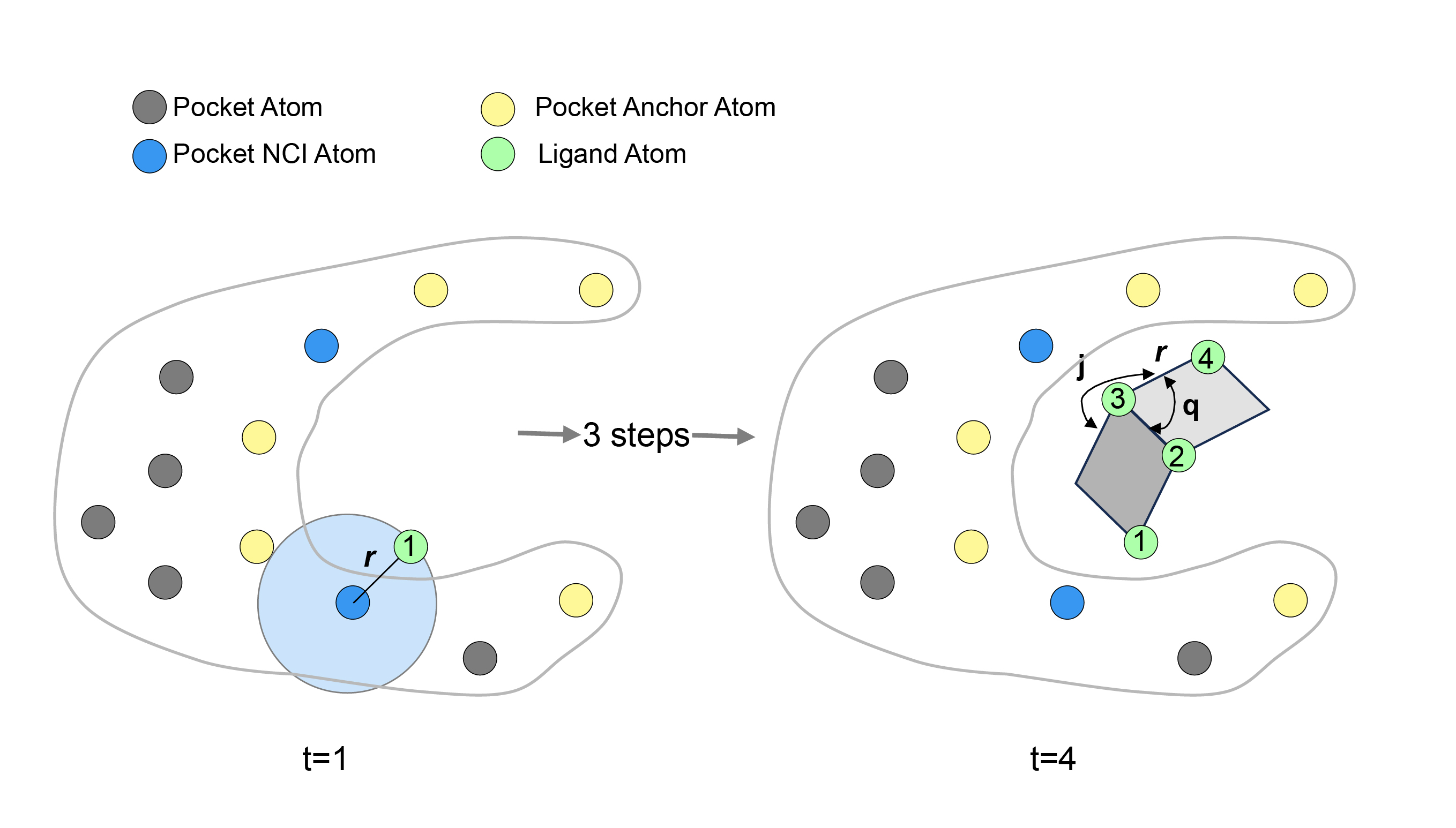}
  \caption{Intuitive visualization of the 3D molecule generation process. In Step 1, based on the precomputed pocket NCI information, we select an NCI as the starting position. Within a radius $r$, we select the position with the highest global coordinate probability. In each subsequent step, we predict the local coordinates $r$, $\theta$, and $\phi$, and combine them with the global coordinates to determine the final position.}
\label{fig:3DMGGen}
\end{figure}

Here, we present the process of generating the final 3D molecule, as shown in Extended Data Fig.\ref{fig:3DMGGen}.

\subsubsection{Atom Generation} \label{ssub:atom_generation}
First, the NCI/Anchor prediction model predicts the NCI sites and anchor atoms.  We then sampled one NCI site from these predictions, and generated the first ligand atom. This atom was positioned at the global coordinates $(x, y, z)$ with the highest predicted joint probability, provided it lies within a 4.5 $\AA$ radius of the sampled NCI site. The subsequent atomic positions were generated iteratively. For each step $i$, $D_{\text{2D}}$ predicts the $(i+1)^\text{th}$ FSMILES token and local coordinates $(r, \theta, \phi)$. Based on $(i+1)^\text{th}$ FSMILES token, we identified the indices of the root1, root2, and root3 atoms. The 3D global coordinates decoder $D_{\text{3D}}$ was employed to predict the probability distribution of the $(i+1)^\text{th}$ 3D coordinates $(x, y, z)$ by incorporating information from the $(i+1)^\text{th}$ FSMILES token and the local coordinates.

\textbf{Leveraging Local and Global Coordinates}. Within a molecule, bond lengths and angles are largely fixed, and FSMILES fragments are consistently rigid and replicable. As a result, the prediction of local spatial positions will get easier by using local coordinates, which include bond length, bond angle, and dihedral angle. In contrast, the global coordinates offer a robust global 3D perception, which is essential for assessing the overall structural context. 

We proposed a fusion method that combines the local and global coordinates. Particularly, this method defines a flexible search space around the predicted local coordinates, then selects the global coordinates with the highest probability. The search space is defined as follows:
\begin{itemize}
    \item $r \pm 0.1 \AA$ (angstroms) for distance.
    \item $\theta \pm 2^\circ$ (degrees) for the angle.  
    \item $\phi \pm 2^\circ$ (degrees) for the dihedral angle.
\end{itemize}

Within this search space, we determined the position with the highest joint probability in the predicted global coordinate distributions. The generation process was repeated to extend the molecular structure progressively. The pseudocode is shown in Supplementary Information Algorithm \ref{alg:3dactiongenaction}.

\subsubsection{Sampling Strategy} \label{ssub:sampling_strategy}
We used $\text{State}(t) = (\text{pocket}, \text{ligand}[t])$ to characterize the generative status at step $t$, where $\text{ligand}[t]$ represents the ligand state after step $t$ is completed. The $\text{Action}(t)$ consists of fragments generated by the encoder/decoder model based on $\text{State}(t)$. Thus, within the framework of this sampling strategy, the encoder/decoder model functions as an action generator under the context of $\text{State}(t)$. When the system adopts a certain $\text{Action}(t)$ under the condition $\text{State}(t) = (\text{pocket}, \text{ligand}[t])$, it moves to the next state $\text{State}(t+1) = (\text{pocket}, \text{ligand}[t+1])$ with a probability of 1, as $P(\text{State}(t+1)| \text{State}(t), \text{Action}(t))= 1$. It is particularly noted that State(0) = (pocket, $\cdot$).

The Reward($t$) is defined to evaluate State($t$) by using two metrics: the model's predictive confidence and the degree of anchor fulfillment. The computation of the model's predictive confidence involves averaging the conditional probabilities of each token involved in Action($t$). The degree of anchor fulfillment was measured by calculating the proportion of anchors that are within 4 $\AA$ of a ligand atom.

To sample at step $t$, the encoder/decoder model utilizes an independent and identically distributed approach based on State($t$) to generate $N$ instances of Action($t$). Then, instances containing atoms with less than 2.5 $\AA$ distance from non-hydrogen pocket atoms were discarded to avoid potential clashes. From the remaining set of Action($t$), we individually summed the normalized model’s confidence and the degree of anchor fulfillment to calculate their respective Rewards. We then retained the top $0.2 \times N$ instances of Action($t$) with the highest Rewards. 

The entire sampling process is executed through a Depth-First Search (DFS) methodology, ensuring a coherent and systematic progression throughout the entire sampling procedure. The pseudocode is shown in Supplementary Information Algorithm \ref{alg:3dsamplegenaction}.

\section{Data Availability}
The evaluation dataset CrossDocked2020 are from the previous study TargetDiff\cite{guan20233d} and is available at their GitHub \href{https://github.com/guanjq/targetdiff}{https://github.com/guanjq/targetdiff}, and the DUD-E\cite{mysinger2012directory}  dataset is a publicly available dataset and is available on our GitHub \href{https://github.com/stonewiseAIDrugDesign/Lingo3DMol}{https://github.com/stonewiseAIDrugDesign/Lingo3DMol}. The PDBbind\cite{renxiao2004pdbbind} dataset is publicly available at \href{http://pdbbind.org.cn/}{http://pdbbind.org.cn/}. The NCI training dataset's NCI label are labeled using an open-source tool, ODDT\cite{Maciej2015oddt}. The protein-ligand complex structures used for model fine-tuning, the generated molecules used for evaluation, and a part of the molecules used for pretraining are accessible via figshare repository\cite{data_for_lingo3dmol}. The full pre-training dataset is a private in-house dataset including molecules sourced from both commercial databases and publicly available databases. Due to contractual obligations with the commercial database vendors, we are unable to share the full pretraining dataset publicly. Nonetheless, we are pleased to offer partial data, specifically 1.4 million molecules, which were obtained from publicly available databases. To request access to additional data, we kindly ask interested researchers to contact the corresponding authors with a proposal outlining their non-commercial research intentions. Upon receipt of a research proposal, we will review it on a case-by-case basis and work towards finding a suitable solution that adheres to the contractual obligations while promoting scientific progress.

\section{Code Availability}
The source code for inference and model architecture is publicly available on GitHub. The pre-training, fine-tuning, and NCI model checkpoints are also available on our GitHub \href{https://github.com/stonewiseAIDrugDesign/Lingo3DMol}{https://github.com/stonewiseAIDrugDesign/Lingo3DMol} and figshare repository\cite{code_for_lingo3dmol}. The model is also available as an online service at \href{https://sw3dmg.stonewise.cn}{https://sw3dmg.stonewise.cn}.
\newline
\newline
\textbf{Acknowledgements}
\newline This study was funded by the National Key R\&D Program of China (grant number 2022YFF1203004 received by BH). This work was also supported by the Beijing Municipal Science and Technology Commission (grant number Z211100003521001 received by JZ and WZ).
\newline
\newline
\textbf{Author Contributions}
\newline WZ and BH conceived the study. WZ and JZ provided instructions for AI modeling. BH and HW provided instructions on evaluation framework. WF, LW, ZL, YZ, RB, HW, and JD developed the model. LW and RB prepared evaluation data. WP supported molecular docking tests.
\newline
\newline
\textbf{Competing Interests}
\newline The authors declare no competing interests.

\afterpage{\clearpage}
\begin{table}[h]
\caption{Comparison of generated drug-like molecules on DUD-E targets (N=101).}
\begin{tabular}{@{}lcccc@{}}
\toprule
\multicolumn{1}{l}{} &
  \multicolumn{1}{l}{\textbf{Random Test}} &
  \multicolumn{1}{l}{\textbf{Pocket2Mol}} &
  \multicolumn{1}{l}{\textbf{TargetDiff}} &
  \multicolumn{1}{l}{\textbf{Lingo3DMol (ours)}} \\ \midrule
\# of molecules generated               & 100,195      & 98,332 & 92,727 & 100,428         \\
\midrule
Mean QED ($\uparrow$)                            & \textbf{0.69}         & 0.46   & 0.50    & 0.53            \\
\midrule
Mean SAS ($\downarrow$)                            & \textbf{2.6}          & 4.0      & 4.9    & 3.3             \\
\midrule
\# of drug-like molecules               & 98,432       & 59,936 & 45,210 & \textbf{82,637} \\
(\% of total generated   molecules) ($\uparrow$) & 98\%        & 61\%  & 49\%  & \textbf{82\%}  \\
\midrule
\multicolumn{5}{l}{\textit{Below comparison only involves drug-like molecules}}            \\
\midrule
\qquad Mean molecular weight                   & 370          & 386    & 299    & 348             \\
\midrule
\qquad ECFP\_TS$>$0.5 ($\uparrow$)           & 17\%         & 8\%   & 3\%    & \textbf{33\%}   \\
\midrule
\qquad Mean min-in-place GlideSP   score ($\downarrow$)   & N/A          & -6.7   & -6.2   & \textbf{-6.8}   \\
\midrule
\qquad Mean GlideSP redocking score   ($\downarrow$)      & -6.4         & -7.5   & -7.0     & \textbf{-7.8}   \\
\midrule
\qquad Mean QED ($\uparrow$)                            & \textbf{0.70} & 0.56    & 0.60    & 0.59             \\
\midrule
\qquad Mean SAS ($\downarrow$)                       & \textbf{2.6} & 3.5    & 4.0      & 3.1             \\
\midrule
\qquad Diversity ($\uparrow$)                           & 0.85         & 0.84   & \textbf{0.88}   & 0.82            \\
\midrule
\qquad Dice ($\uparrow$)                           & 0.21         & 0.24   & \textbf{0.28}   & 0.25            \\
\midrule
\qquad \makecell[l]{Mean RMSD \\ \textit{vs}.\\ low-energy conformer ($\AA$,$\downarrow$)} & 4.0 & 1.1 & 1.1 & \textbf{0.9} \\ 
\bottomrule
\end{tabular}
{\raggedright Note: For each method, we generated approximately 1000 molecules per target. To determine the drug-likeness and inclusion in the comparison, we considered molecules with a QED score greater than or equal to 0.3 and a SAS less than or equal to 5. The metric "ECFP\_TS$>$0.5" represents the percentage of targets with generated compounds that are similar to active compounds based on the Tanimoto similarity of ECFP4\cite{tanimoto1958elementary}. The min-in-place GlideSP score and GlideSP redocking score were calculated using the Glide software. The RMSD value indicates the differences between the generated conformers and the low-energy conformers generated using ConfGen\cite{watts2010confgen}. As for the random set, we randomly selected 1000 molecules from our in-house commercial library for each target. Since there are no "generated conformers" for the random test molecules, the RMSD in this case represents the differences between the docked conformer and the low-energy conformer. More details of molecular weight distribution could be found in Extended Data Fig.\ref{fig:Adist}. Diversity reflects the average pair-wise Tanimoto similarity of molecules generated for the same target. Dice score was defined as the ratio of "intersection over union" between the voxelized representations of the reference compounds observed in the crystal structure (i.e. PDB ID) and the generated molecules. To calculate Dice score, we created a grid with points at 0.5 $\AA$ intervals to cover both molecules. Each grid point was evaluated to determine if it fell within the 1.2 times of covalent radius (referred as testing radius) of any atom in either molecule. Grid points within the testing radius of atoms in both molecules were considered as intersected points, while grid points within the testing radius of any atom in either molecule were considered as union points. The Dice score, calculated as the ratio of intersected points over union points, ranges from 0 to 1, with a value of 0 indicating no similarity or overlap between the molecules. Bold face indicates best performance.}
\label{tab:DUD-E}
\end{table}

\begin{table}[h]
\caption{Comparison of generated drug-like molecules involved in ablation studies.}
\begin{tabular}{@{}lccc@{}}
\toprule
   & \makecell{\textbf{Lingo3DMol} \\ \textbf{(standard)}} 
   & \makecell{\textbf{Lingo3DMol} \\ \textbf{(with random NCI})}
   & \makecell{\textbf{Lingo3DMol} \\ \textbf{(without pre-training})}
   \\ 
\midrule
\# of molecules generated                                   & 100,428  & 99,170      & 23,982       \\
\midrule
Mean QED ($\uparrow$)                                             & \textbf{0.53}     & 0.35        & 0.46         \\
\midrule
Mean SAS ($\downarrow$)                                             & \textbf{3.3}     & 3.5         & 3.4          \\
\midrule
\# of drug-like molecules                                   & \textbf{82,637}    & 46,502      & 16,966       \\

(\% of total generated molecules) ($\uparrow$)                       & \textbf{82\%}   & 47\%        & 71\%         \\
\midrule
\multicolumn{4}{l}{\textit{Below comparison only involves drug-like molecules}}  \\
\midrule
\qquad Mean molecular weight                                       & 348     & 424         & 345          \\
\midrule
\qquad ECFP\_TS$>$0.5 ($\uparrow$)                               & \textbf{33\%}       & 6\%        & 3\%          \\
\midrule
\qquad Mean min-in-place GlideSP score ($\downarrow$)                         & \textbf{-6.8}     & -5.8        & -4.9         \\
\midrule
\qquad Mean GlideSP redocking score ($\downarrow$)                            & \textbf{-7.8}      & -7.2        & -6.9         \\
\midrule
\qquad Mean QED ($\uparrow$)                                                & \textbf{0.59}      & 0.51        & 0.56         \\
\midrule
\qquad Mean SAS ($\downarrow$)                                           & \textbf{3.1}       & 3.3         & \textbf{3.1}          \\
\midrule
\qquad Diversity ($\uparrow$)                                               & 0.82           & \textbf{0.83}            & 0.70             \\
\midrule
\qquad Dice ($\uparrow$)                           & \textbf{0.25}   & 0.15   & 0.13            \\
\midrule
\qquad \makecell[l]{Mean RMSD \\ \textit{vs}.\\ low-energy conformer ($\AA$,$\downarrow$)}                      & \textbf{0.9}       & 1.3         & 1.0 \\
\bottomrule
\end{tabular}
{\raggedright Note: For Lingo3DMol stardard and Lingo3DMol with random NCI, we generated approximately 1000 molecules per target; for Lingo3DMol without pre-training, with the same computational resources and time constraints, molecules can only be generated on 73 pockets. Bold face indicates best performance.}
\label{tab:ablation_test}
\end{table}

\newpage

\begin{figure}[H]
  \centering
  \includegraphics[width=1.0\textwidth]{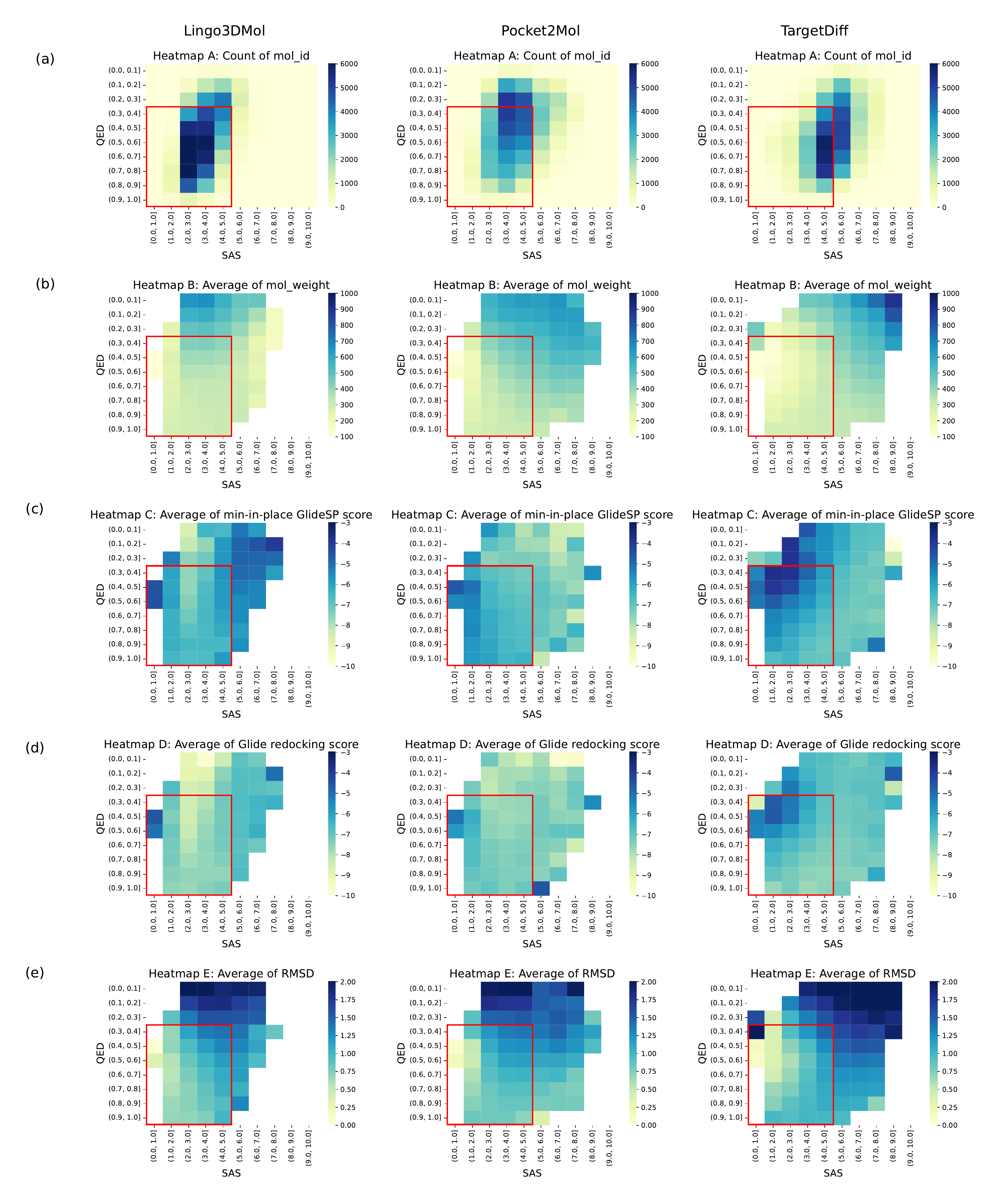}
  \caption{Distributions of molecules generated by Lingo3DMol, Pocket2Mol, and TargetDiff on DUD-E targets (N=101). The drug-like region with QED$>$=0.3 and SAS$<$=5 is indicated with red boxes. Heatmaps in panel (a), (b), (c), (d), and (e) display the distribution of key properties for generated molecules, including count, molecular weight, minimum in-place GlideSP score, GlideSP redocking score, and RMSD vs. low-energy conformer, respectively. These distributions are depicted along SAS and QED.}
  \label{fig:mol_dist}
\end{figure}

\afterpage{\clearpage}
\textbf{Fig.\ref{fig:mol_dist}} Distributions of molecules generated by Lingo3DMol, Pocket2Mol, and TargetDiff on DUD-E targets (N=101). The drug-like region with QED$>$=0.3 and SAS$<$=5 is indicated with red boxes. Heatmaps in panel (a), (b), (c), (d), and (e) display the distribution of key properties for generated molecules, including count, molecular weight, minimum in-place GlideSP score, GlideSP redocking score, and RMSD vs. low-energy conformer, respectively. These distributions are depicted along SAS and QED.

\begin{figure}[H]
  \centering
  \includegraphics[width=1.0\textwidth]{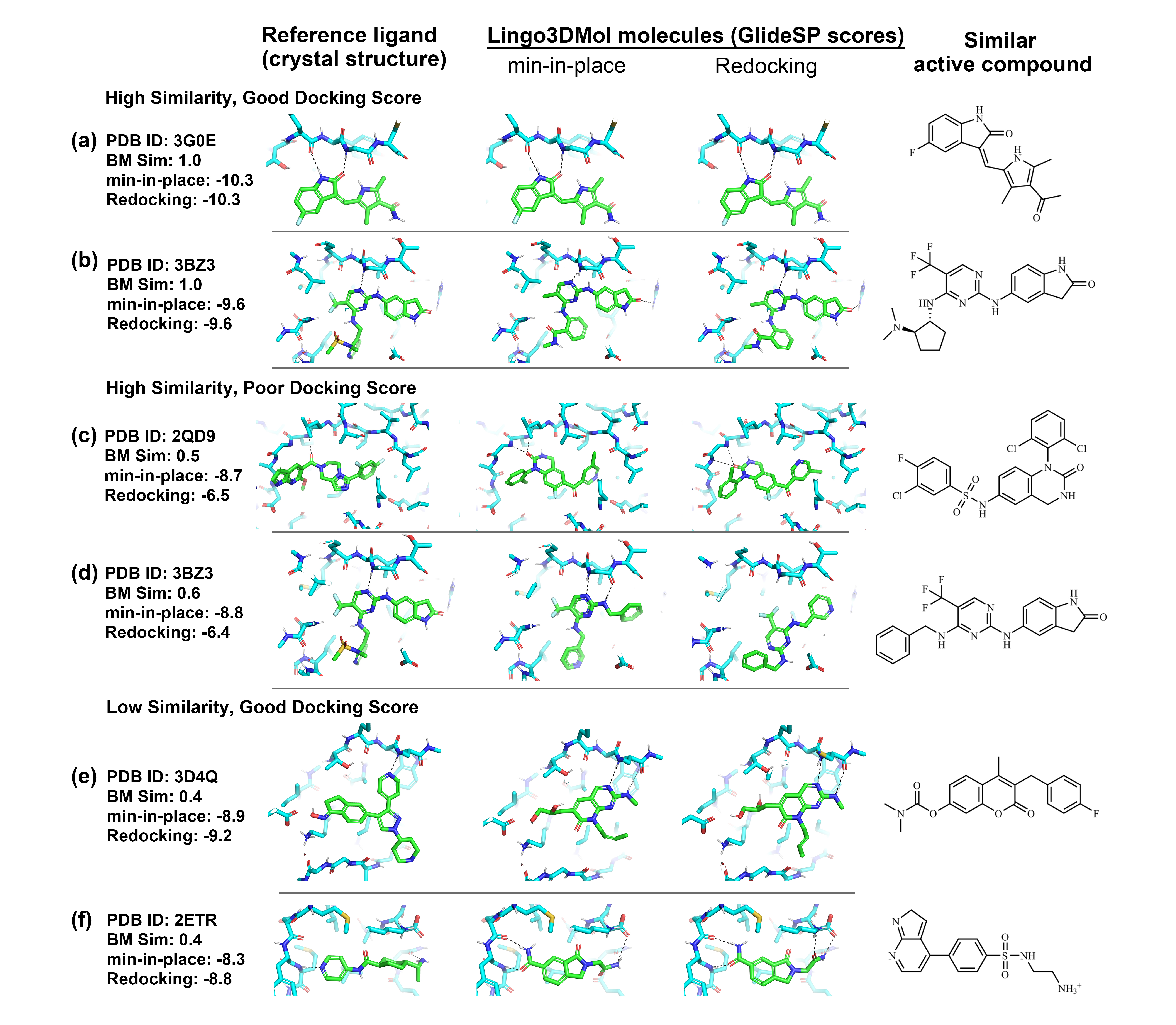}
  \caption{Case study of generated molecules involving 3D binding mode and 2D similarity with active compounds. The reference binding mode is based on the crystal structure from PDB. The Lingo3DMol conformation represents the generated conformation, while the GlideSP redocking conformation was obtained by docking the generated compound into the pocket using Glide with specific parameters (DOCKING\_METHOD: confgen and PRECISION: SP). The most similar known active compound to the generated molecule is also displayed, noting that it may not necessarily be the reference compound observed in the crystal structure. The information provided includes the PDB ID for the reference binding mode, the Tanimoto similarity between the BM scaffold of generated molecule and its most similar active compound, and the GlideSP redocking score.}
  \label{fig:my_fig}
\end{figure}

\textbf{Fig.\ref{fig:my_fig}} Case study of generated molecules involving 3D binding mode and 2D similarity with active compounds. The reference binding mode is based on the crystal structure from PDB. The Lingo3DMol conformation represents the generated conformation, while the GlideSP redocking conformation was obtained by docking the generated compound into the pocket using Glide with specific parameters (docking\_method: confgen and precision: sp). The most similar known active compound to the generated molecule is also displayed, noting that it may not necessarily be the reference compound observed in the crystal structure. The information provided includes the PDB ID for the reference binding mode, the Tanimoto similarity between the BM scaffold of generated molecule and its most similar active compound, and the GlideSP redocking score.

\begin{figure}[H]
\centering
\includegraphics[width=1.0\textwidth]{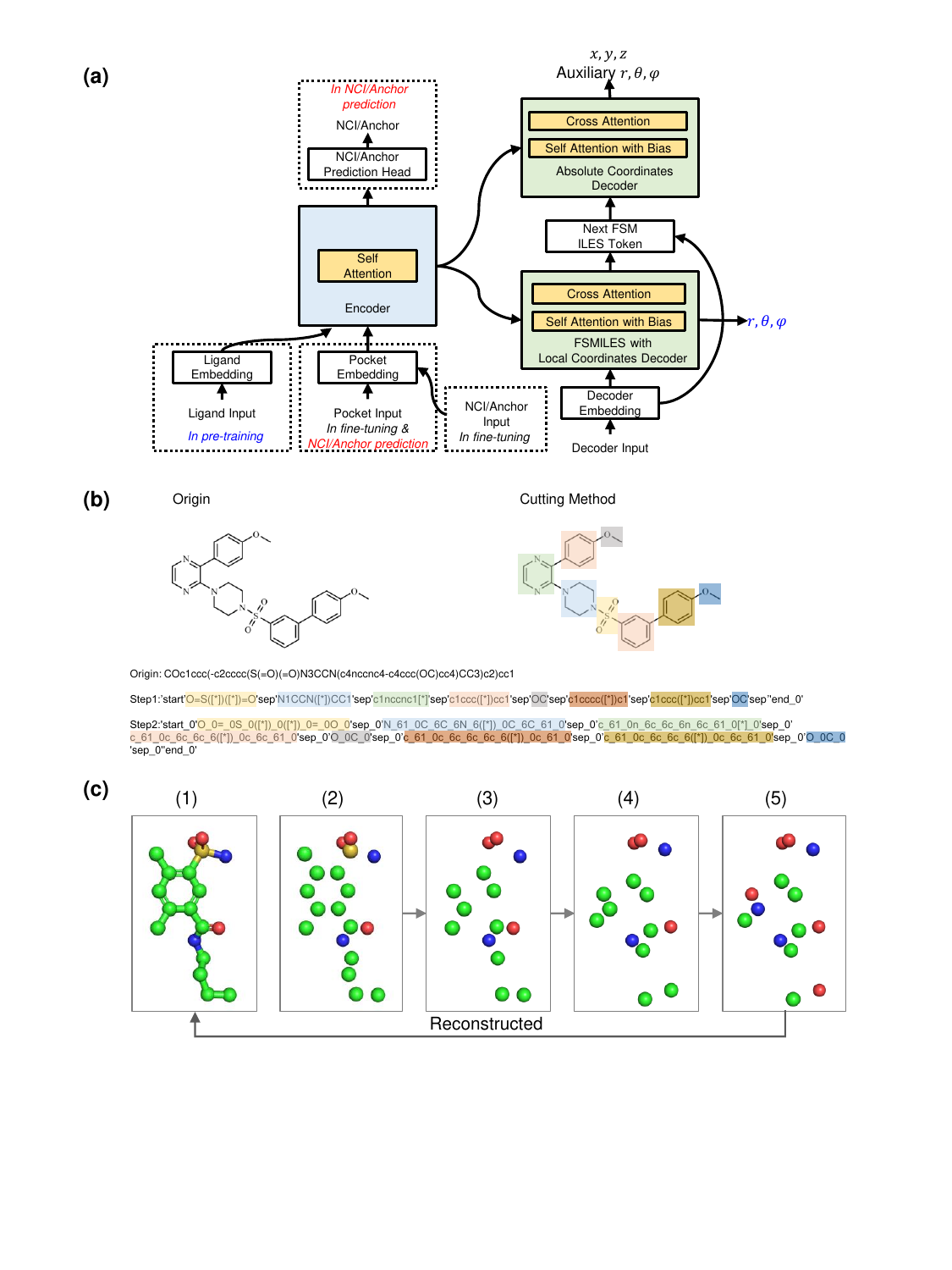}
\caption{Overview of Lingo3DMol model development. (a) Lingo3DMol architecture. Three separate models are included: the pre-training model, the fine-tuning model, and the NCI/Anchor prediction model. These models share the same architecture with slightly different inputs and outputs. (b) Illustration of FSMILES construction. The same color corresponds to the same fragment. (c) Illustration of pre-training perturbation strategy. Step 1: original molecular state; step2: removal of edge information during pre-training; step 3: perturbation of the molecular structure by randomly deleting 25\% of the atoms; step 4: perturbation of the coordinates using a uniform distribution within the range [-0.5 $\AA$, 0.5 $\AA$]; step 5: perturbation of 25\% of the carbon element types. These perturbations are applied in no particular order,and the pre-training task aims to restore the molecular structure from step 5 to step 1.}
\label{fig:3DMG}
\end{figure}

\textbf{Fig.\ref{fig:3DMG}} Overview of Lingo3DMol model development. (a) Lingo3DMol architecture. Three separate models are included: the pre-training model, the fine-tuning model, and the NCI/Anchor prediction model. These models share the same architecture with slightly different inputs and outputs. (b) Illustration of FSMILES construction. The same color corresponds to the same fragment. (c) Illustration of pre-training perturbation strategy. Step 1: original molecular state; step2: removal of edge information during pre-training; step 3: perturbation of the molecular structure by randomly deleting 25\% of the atoms; step 4: perturbation of the coordinates using a uniform distribution within the range [-0.5 $\AA$, 0.5 $\AA$]; step 5: perturbation of 25\% of the carbon element types. These perturbations are applied in no particular order,and the pre-training task aims to restore the molecular structure from step 5 to step 1.

\newpage
\bigskip
\bibliography{sn-bibliography}

\newpage
\afterpage{\clearpage}

\newpage
\setcounter{page}{1}
\afterpage{\clearpage}
\subsection*{Extended Data}
\beginextendeddata
\begin{table}[h]
\caption{Comparison of bond lengths between the reference molecules and the generated molecules.}
\begin{tabular*}{\linewidth}{@{}@{\extracolsep{\fill}}lrrrrrrrr@{}}
\toprule
\multicolumn{1}{c}{\multirow{2}{*}{Bond}} & \multicolumn{2}{c}{Reference}                      & \multicolumn{2}{c}{Pocket2Mol}                     & \multicolumn{2}{c}{TargetDiff}                     & \multicolumn{2}{l}{Lingo3DMol}                     \\ \cmidrule(l){2-9} 
\multicolumn{1}{c}{}                      & \multicolumn{1}{l}{Mean} & \multicolumn{1}{l}{Std} & \multicolumn{1}{l}{Mean} & \multicolumn{1}{l}{Std} & \multicolumn{1}{l}{Mean} & \multicolumn{1}{l}{Std} & \multicolumn{1}{l}{Mean} & \multicolumn{1}{l}{Std} \\
\midrule
C-C                                       & 1.52                     & 0.05                    & 1.45                     & 0.11                    & 1.48                     & 0.08                    & 1.51                     & 0.10                     \\
\midrule
C=C                                       & 1.39                     & 0.07                    & 1.37                     & 0.10                     & 1.39                     & 0.07                    & 1.40                      & 0.12                    \\
\midrule
C:C                                       & 1.41                     & 0.04                    & 1.39                     & 0.09                    & 1.39                     & 0.03                    & 1.40                      & 0.07                    \\
\midrule
C-N                                       & 1.42                     & 0.07                    & 1.39                     & 0.10                     & 1.41                     & 0.07                    & 1.46                     & 0.23                    \\
\midrule
C=N                                       & 1.34                     & 0.05                    & 1.34                     & 0.11                    & 1.35                     & 0.06                    & 1.38                     & 0.23                    \\
\midrule
C:N                                       & 1.36                     & 0.03                    & 1.35                     & 0.10                     & 1.36                     & 0.04                    & 1.36                     & 0.07                    \\
\midrule
C-O                                       & 1.41                     & 0.05                    & 1.38                     & 0.09                    & 1.40                      & 0.07                    & 1.40                      & 0.07                    \\
\midrule
C=O                                       & 1.24                     & 0.04                    & 1.26                     & 0.09                    & 1.28                     & 0.05                    & 1.23                     & 0.07                    \\
\midrule
C:O                                       & 1.45                     & 0.02                    & 1.39                     & 0.11                    & 1.41                     & 0.05                    & 1.38                     & 0.04                    \\ 
\bottomrule
\end{tabular*}
\label{tab:bond_lengths}
\end{table}

\begin{table}[h]
\caption{Comparison of the ring size distribution in molecules generated by different methods.}
\begin{tabular*}{\linewidth}{@{}@{\extracolsep{\fill}}rrrrr@{}}
\toprule
\multicolumn{1}{l}{Ring Size} & \multicolumn{1}{l}{Reference} & \multicolumn{1}{l}{Pocket2Mol} & \multicolumn{1}{l}{TargetDiff} & \multicolumn{1}{l}{Lingo3DMol} \\ \midrule
3   & 1.62\% & 0.12\%  & 0.00\%  & 0.18\%  \\
\midrule
4   & 0.00\% & 0.02\%  & 2.70\%  & 1.28\%  \\
\midrule
5   & 29.55\% & 16.26\% & 29.71\% & 34.71\% \\
\midrule
6   & 65.99\% & 79.83\% & 48.96\% & 63.45\% \\
\midrule
7   & 0.81\% & 2.59\%  & 11.70\% & 0.23\%  \\
\midrule
8   & 0.00\% & 0.34\%  & 2.59\%  & 0.11\%  \\
\midrule
9   & 0.00\% & 0.12\%  & 0.85\%  & 0.02\%  \\
\midrule
10+ & 2.02\% & 0.72\%  & 3.48\%  & 0.01\%  \\ 
\bottomrule
\end{tabular*}
\label{tab:ring_size}
\end{table}

\begin{table}[h]
\caption{In-place GlideSP score analysis for DUD-E targets (N=101).}
\begin{tabular}{lrrr}
\toprule
\textbf{} & \multicolumn{1}{l}{\textbf{Lingo3DMol}} & \multicolumn{1}{l}{\textbf{Pocket2Mol}} & \multicolumn{1}{l}{\textbf{TargetDiff}} \\ \midrule
\# of molecules generated                                                       & 100,428 & 98,332 & 92,727 \\
\midrule
\% of molecules with positive in-place GlideSP score                            & 84\%   & 87\%   & 60\%   \\
\midrule
\makecell[l]{Mean in-place GlideSP score \\ (All molecules)}                                   & 8,450  & 8,744  & 6,127  \\
\midrule
\makecell[l]{Mean in-place GlideSP score \\ (Molecules with positive in-place GlideSP score)} & 9,967  & 9,980  & 9,884  \\
\midrule
\makecell[l]{Mean in-place GlideSP score \\ (Molecules with Negative in-place GlideSP score)} & -4.9   & -5.1   & -5.2   \\ \bottomrule
\end{tabular}
\label{tab:glidesp_anlysis}
\end{table}

\begin{table}[h]
\caption{Inference time for Lingo3DMol, Pocket2Mol and TargetDiff.}
\begin{tabular*}{\linewidth}{@{}@{\extracolsep{\fill}}lccc@{}}
\toprule
   & Lingo3DMol  & Pocket2Mol  & TargetDiff\\   
\midrule
Running time (s, $\downarrow$) & \textbf{874$\pm$401} & 962$\pm$622 & 1327$\pm$405 \\ \bottomrule
\end{tabular*}
{\raggedright Note: We randomly selected 10 targets from DUD-E and recorded the time taken to generate 100 valid molecules for each target using an NVIDIA Tesla V100.}
\label{tab:running_time}
\end{table}

\begin{figure}[!h]
  \centering
  \includegraphics[width=1.0\linewidth]{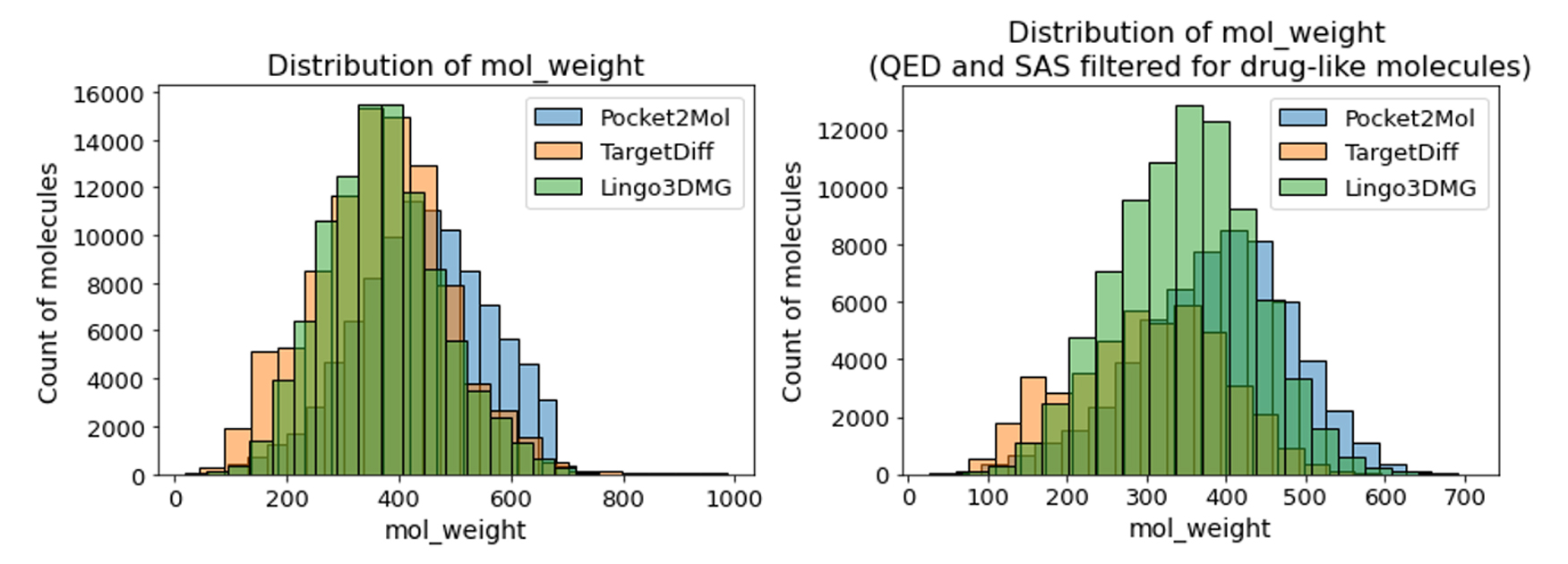}
  \caption{Distribution of molecular weight for molecules generated by different methods. This figure is provided as supplementary information for Table \ref{tab:DUD-E}.}
\label{fig:Adist}
\end{figure}

\newpage
\afterpage{\clearpage}
\addtocontents{toc}{\setcounter{tocdepth}{2}}
\section*{Supplementary Information}
{
\titlecontents{subsection}[2.0em]{}{\thecontentslabel\quad}{\hspace{4.0em}}{\titlerule*[0.75pc]{.}\contentspage}
\tableofcontents
\thispagestyle{empty}
\setcounter{page}{0}
\newpage
}
\beginsupplement

\subsection{Validation Method for Pre-training Effect}\label{pre_train_supl}
The method employed in this study involved generating two sets of molecules, Set A and Set B, on the DUD-E targets using different models. Set A was generated by the model with pre-training, while Set B was generated by the model without pre-training. To compare the similarity of Set A and Set B to the pre-training set, a chemical space based distribution analysis was conducted. Specifically, the molecules in Set A, Set B, and the pre-training set were projected onto a two-dimensional chemical space, represented by a 120X120 lattice self-organizing map (SOM). The methodologies for constructing the SOM-based chemical space and projecting molecules onto this space were previously described in our earlier study\cite{wang2022pocket}. Subsequently,  total variation divergence and KL divergence were calculated to compare the distribution similarity of Set A and Set B to the pre-training set. The results, presented in Table \ref{tab:pretrain_compare}, indicate that Set A exhibits a higher degree of similarity to the pre-training set compared to Set B.

\begin{equation}
D_{TV} (P\|Q) = \frac{1}{2} \sum\limits_{i} \left|P_i - Q_i\right|,
\end{equation}

\begin{equation}
D_{KL} \left(P\|Q\right) = E_{x\sim P} \text{log}  \left[\frac{P\left(x\right)}{Q\left(x\right)}\right].
\end{equation}

\begin{table}[tbhp]
    \centering
    \caption{Distribution similarity of Set A and Set B to the pre-training set.}
    \label{tab:pretrain_compare}
    \begin{tabular*}{\linewidth}{@{}@{\extracolsep{\fill}}cccccc@{}}
         \toprule
               &\makecell{\textbf{Total variation divergence}}   &\textbf{KL divergence} \\
         \midrule
         \makecell{Molecules generated with pre-training (Set A)  \\ \textit{vs}.\\  Molecules in pre-training set} &0.51 &2.33 \\
         \midrule
         \makecell{Molecules generated without pre-training (Set B) \\ \textit{vs}.\\ Molecules in pre-training set} &0.60 &3.72 \\
         \bottomrule
    \end{tabular*}
{\raggedright Note: Additional details about the molecules in Set A and Set B can be found in Table \ref{tab:ablation_test}.}
\end{table}

\subsection{Algorithms}
Here, we demonstrated three algorithms: the algorithm for atom generation (Supplementary Information Algorithm 
 \ref{alg:3dactiongenaction}), the algorithm for molecular generation sampling strategy (Supplementary Information Algorithm 
 \ref{alg:3dsamplegenaction}), and the algorithm for model training (Supplementary Information Algorithm \ref{alg:training_process}).

\begin{algorithm}[H]
\caption{Model training.} \label{alg:training_process}
\begin{algorithmic}[1] 
\State Procedure Generate ligand FSMILES

\State \hspace{1em}input the molecule SDF file 
\State \hspace{1em}Find cuttable single bonds that meets the following criteria: (1) not in a ring, (2) not connected to hydrogen atoms, (3) at least one end of the single bond is attached to a ring
\State \hspace{1em}Divide the molecule into fragments by breaking the cuttable single bonds.
\State \hspace{1em}Concatenate the SMILES of each fragment using a symbol $sep$ to obtain FSMILESraw

\For{each token T in FSMILESraw}
    \State \hspace{1em} if T represents an atom and T is in a ring do 
    \State \hspace{2em} update T by concatenating the size information of the atom's associated ring with T.
    \State \hspace{1em} else 
    \State \hspace{2em} update T by concatenating 0 with T
    \State \hspace{1em} end if 
\EndFor
\State \hspace{1em}FSMILESraw is the FSMILES
~\\
\State Procedure Generate pocket representation
\State \hspace{1em} input the pocket PDB file

\For{each heavy atom in the protein pocket}
    \State \hspace{1em}Get the properties of the atom including atom type, is\_aromatic, has\_hydrogen, hba (hydrogen bond acceptor) or hbd (hydrogen bond donor)
    \State \hspace{1em}Get the aromatic centers
    \State \hspace{1em}Choose the atoms within 4 $\AA$ distance to the ligand, as anchors
    \State \hspace{1em}Randomly choose one of NCI atoms as the starting point for generation
    \State \hspace{1em}Encode properties, aromatic centers and anchors into pocket representation
\EndFor
~\\

\State epoch num = 200 
\State batch size = 50 
\State FSMILES token length = 100
\State pocket representation length = 500
\State shuffle the data in dataloader 

\For{each epoch}
\For{each batch} 
\For{each complex data}
\State Rotate the complex randomly
\State Generate the ligand FSMILES
\State Get the ligand global and local coordinates, root node, symbol type
\State Generate the pocket representation. 
\EndFor 
\State Use teacher forcing to predict the FSMILES next token and coordinates autoregressively
\State Calculate the token loss, global and local coordinates losses. 
Sum the losses as total loss
\State Backpropagate the total loss and use Adam optimizer to update the model parameters
\EndFor 
\EndFor
\end{algorithmic}
\end{algorithm}

\begin{algorithm}[H]
\caption{Molecule generation process.} \label{alg:3dactiongenaction}
\begin{algorithmic}[1] 
\State Initialize first atom position(this procedure is executed solely for the case of Action(0)):

\State \hspace{1em}Randomly select an NCI site from the list of predicted NCI sites provided by the NCI predictor
\State \hspace{1em}Choose the highest predicted joint probability for initial $(x, y, z)$ within radius $r$

\For{each generation step $i$ in Action(t)}
    \State Predicts the $(i+1)^\text{th}$ FSMILES token
    \State Identify the indices of the  $(i+1)^\text{th}$'s root1, root2, and root3 atoms based on FSMILES codes
    \State Predicts local coordinates $(r, \theta, \phi)$ based on roots features.
    \State $D_{\text{3D}}$ predicts the $(i+1)^\text{th}$ 3D coordinates $(x, y, z)$ using the FSMILES token
    \State Define search space around predicted local coordinates for final atom coordinates:
        \State \hspace{1em} Distance: $r \pm 0.1 \AA$
        \State \hspace{1em} Angle: $\theta \pm 2^\circ$
        \State \hspace{1em} Dihedral Angle: $\phi \pm 2^\circ$
    \State Find global coordinate with highest joint probability within search space
    \State Set final predicted location for the current generation step
    \State if the $i+1$'s FSMILES token is $sep$ then
    \State \hspace{1em} break
    \State end if 
\EndFor
\end{algorithmic}
\end{algorithm}

\begin{algorithm}[H]
\caption{Sampling strategy for molecule generation using DFS approach.} \label{alg:3dsamplegenaction}
\begin{algorithmic}[1] 
\State Generate N instances of Action(0) under State(0) using the encoder/decoder model, for each Action(0), transition to State(1)
\State Perform clashes detection for all State(1) instances. 
\State Calculate the rewards for State(1) instances that pass the clashes detection.
\State Retain a maximum of $0.2\times N$ State(1) instances with the highest rewards. 
\State Push the selected State(1) instances into a stack in ascending order based on their rewards.
\For{Stack is not empty}
    \State Pop the element currently at the top of the stack and denote it as State($t$).
    \State Generate $N$ instances of Action($t$) under State($t$) using the encoder/decoder model, for each Action($t$), transition to State($t+1$)
    \State Perform clashes detection for all State($t+1$) instances. 
    \State Calculate the rewards for State($t+1$) instances that pass the clashes detection.
    \State Retain a maximum of $0.2\times N$ State($t+1$) instances with the highest rewards. 
    \State Push the selected State($t+1$) instances into the stack in ascending order based on their rewards.
\EndFor
\end{algorithmic}
\end{algorithm}

\subsection{Running Time}
In terms of generation speed, Lingo3DMol is faster than benchmark methods. We randomly selected 10 targets from DUD-E and recorded the time taken to generate 100 valid molecules for each target using an NVIDIA Tesla V100, the results are displayed in Extended Data Table \ref{tab:running_time}.

\subsection{NCI/Anchor Prediction Model Analysis}
We trained the NCI/Anchor Prediction Model on our labeled PDBbind2020\cite{renxiao2004pdbbind} dataset and evaluated its precision and recall on the DUD-E targets. The results in Table \ref{tab:nci_effect} show that the precision for NCI atoms on pocket is 0.629, and the recall is 0.622. For anchor atoms on pocket, the precision is 0.739 and the recall is 0.756.
\begin{table}[tbhp]
\centering
\caption{Precision/Recall effect of the NCI model on the DUD-E targets (N=101).}
\label{tab:my_label}
\begin{tabular*}{\linewidth}{@{}@{\extracolsep{\fill}}cccccc@{}}
     \toprule
           &\makecell{\textbf{Precision}}   &\textbf{Recall} \\
     \midrule
     \makecell{NCI atoms on pocket} &0.629 &0.622 \\
     \midrule
     \makecell{Anchor atoms on pocket} &0.739 &0.756 \\
     \bottomrule
\end{tabular*}
\label{tab:nci_effect}
\end{table}

\subsection{Encoder/Decoder Backbone}
The encoder and decoder use the standard Transformer\cite{vaswani2017attention} architecture with attention bias in decoder.

\textbf{Encoder}. The encoder has $N$ identical layers, each with multi-head self-attention and a position-wise fully connected feed-forward network. Residual connections and layer normalization are employed, with all outputs having dimension $d_{\text{model}}$\cite{vaswani2017attention}.

\textbf{Decoder}. The decoder also has $N$ identical layers, with an extra multi-head cross-attention sub-layer for the encoder's output. Residual connections and layer normalization are used, and a masked self-attention sub-layer ensures position dependencies\cite{vaswani2017attention}.

\textbf{Self-Attention}.
Let $Q$, $K$, and $V$ represent the query, key, and value matrices, respectively. The original attention scores are calculated as:

\begin{equation}\label{attention}
   A = \text{softmax}\left(\frac{QK^\top}{\sqrt{d_k}}\right)V,
\end{equation}
where $d_k$ is the dimension of the key vectors. 

\subsection{HyperParameters}
We employ a consistent set of hyperparameters for all three stages of our model, namely pre-training, fine-tuning, and NCI/Anchor prediction. These hyperparameters are shown in Table \ref{tab:hyperparams}.

\begin{table}[tbhp]
\centering
\caption{Hyperparameters used for the 3D molecule generation model.}
\begin{tabular*}{\linewidth}{@{}@{\extracolsep{\fill}}lc@{}}
    \toprule
    \textbf{Hyperparameter}  & \textbf{Value} \\ 
    \midrule
    Batch size & 50 \\ 
    \midrule
    Learning rate ($\alpha$) & 1e-4 \\ 
    \midrule
    Hidden dimension & 512 \\ 
    \midrule
    Feed-forward network (FFN) dimension & 2048 \\ 
    \midrule
    Encoder layers & 6 \\ 
    \midrule
    Attention head size & 8 \\ 
    \midrule
    Topological decoder layers & 3 \\ 
    \midrule
    3D global decoder layers & 3 \\ 
    \bottomrule
\end{tabular*}
\label{tab:hyperparams}
\end{table}

\subsection{Random Cases}\label{Random cases}
\begin{figure}[tbhp]
\centering
\includegraphics[width=1\textwidth]{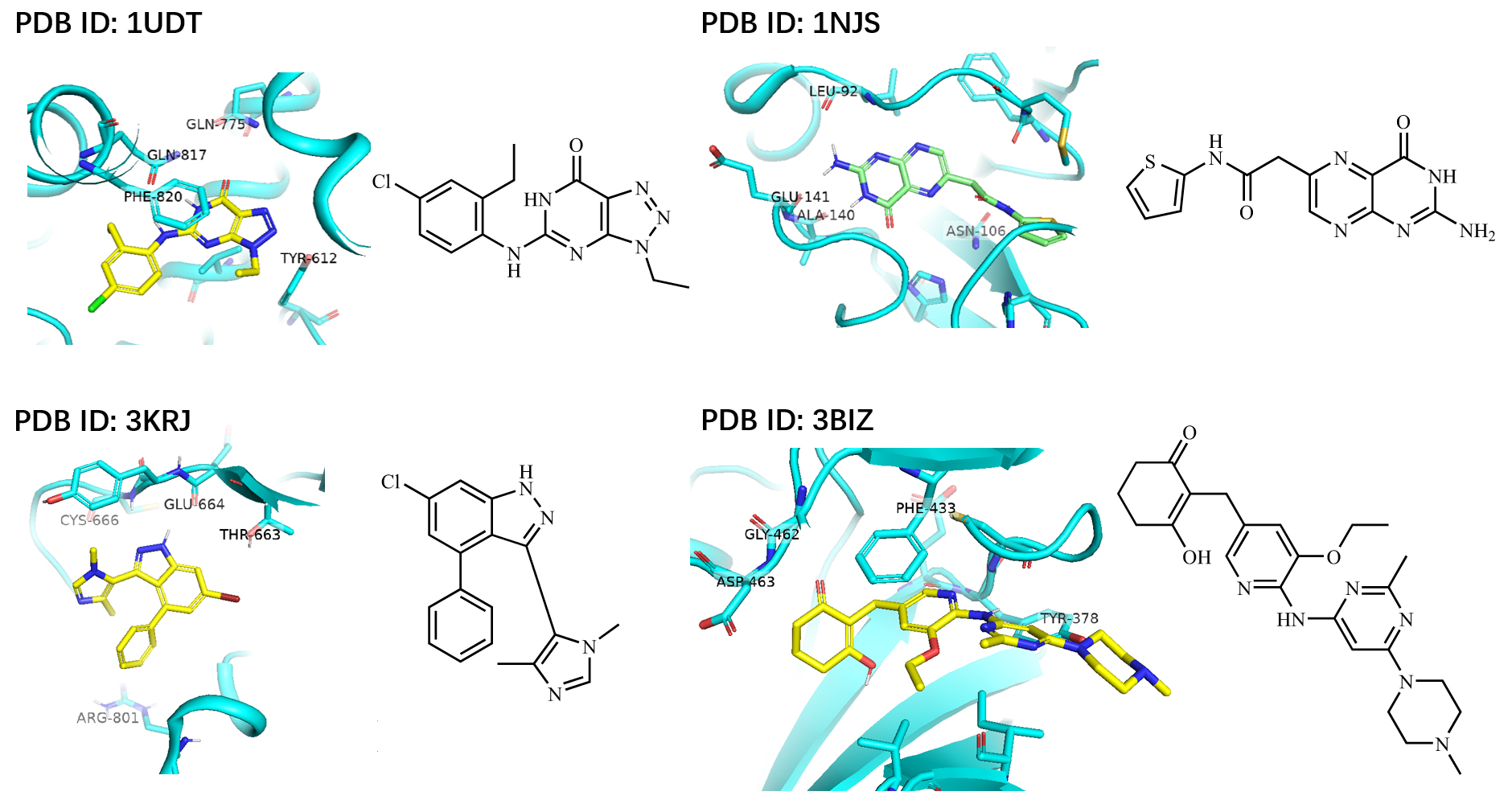}
\caption{Examples of Lingo3DMol generated molecules.}
\label{fig:dude_random_cases}
\end{figure}
\end{document}